\PassOptionsToPackage{capitalize}{cleveref}
\documentclass[11pt]{arxiv/template/style/prism-princeton}
\usepackage{xurl}
\usepackage{arxiv/template/style/prism-cover}
\newif\ifPrismAffiliationNewRow
\newcommand{\affiliationbreak}{\PrismAffiliationNewRowtrue}
\renewcommand{\affiliation}[2][]{%
  \ifdefempty{\PrismCoverAffiliations}{}{%
    \ifPrismAffiliationNewRow
      \gappto\PrismCoverAffiliations{\\}%
      \PrismAffiliationNewRowfalse
    \else
      \gappto\PrismCoverAffiliations{,\space}%
    \fi}%
  \gappto\PrismCoverAffiliations{\PrismCoverMark{#1}#2}}
\patchcmd{\PrismMakeCover}
  {\small\PrismCoverAffiliations\par}
  {\fontsize{9}{11}\selectfont\PrismCoverAffiliations\par}
  {}{\PackageError{prism-author-layout}{Affiliation font patch failed}{Check the registered cover component.}}

\DeclareFontFamily{T1}{SWERefOptimistic}{}
\DeclareFontShape{T1}{SWERefOptimistic}{b}{n}
  {<-> s * [0.88] arxiv/fonts/optimistic}{}
\pdfmapline{+optimistic <arxiv/fonts/Optimistic.ttf <arxiv/fonts/T1-WGL4.enc}
\newcommand{\SWERefCoverBold}{%
  \fontfamily{SWERefOptimistic}\fontseries{b}\fontshape{n}\selectfont}
\patchcmd{\PrismMakeCover}
  {\PrismCoverSemi\fontsize{19}{24}\selectfont\PrismCoverTitle\par}
  {\SWERefCoverBold\fontsize{16.5}{22}\selectfont\PrismCoverTitle\par}
  {}{\PackageError{prism-cover-typography}{Title font patch failed}{Check the registered cover component.}}
\renewcommand{\PrismCoverMetaEntry}[2]{\par{\SWERefCoverBold #1:}\enspace #2}
\patchcmd{\PrismMakeCover}
  {\raggedright\footnotesize\color{PrismCoverInk}\setlength{\parskip}{1pt}}
  {\raggedright\fontsize{9}{12}\selectfont\color{PrismCoverInk}\setlength{\parskip}{0pt}}
  {}{\PackageError{prism-cover-typography}{Metadata font patch failed}{Check the registered cover component.}}

\usepackage{hyperref}
\pdftrailerid{}
\hypersetup{pdfauthor={},pdfcreator={},pdfproducer={}}
\usepackage{url}
\usepackage{mathtools}
\usepackage{amssymb}
\usepackage{bm}
\usepackage{xcolor}
\usepackage{graphicx}
\usepackage{booktabs}
\usepackage{tabularx}
\usepackage{longtable}
\usepackage{pdflscape}
\usepackage{multirow}
\usepackage{makecell}
\usepackage{threeparttable}
\usepackage{float}
\usepackage{wrapfig}
\usepackage{placeins}
\usepackage{needspace}
\usepackage{listings}
\usepackage{algpseudocode}
\usepackage[most]{tcolorbox}
\usepackage{tikz}
\usetikzlibrary{arrows.meta,calc,positioning}
\usepackage{pgfplots}
\usepgfplotslibrary{fillbetween}
\pgfplotsset{compat=1.18}
\tikzset{every picture/.append style={scale=.86,transform shape}}
\usepackage[capitalize,noabbrev]{cleveref}

\definecolor{AreteInk}{HTML}{1B1B1B}
\definecolor{AreteSlate}{HTML}{555555}
\definecolor{AreteNight}{HTML}{1B2340}
\definecolor{AreteBlue}{HTML}{356D9B}
\definecolor{AreteViolet}{HTML}{745EB5}
\definecolor{AreteCyan}{HTML}{347E83}
\definecolor{AreteGold}{HTML}{8A6A35}
\definecolor{AreteCoral}{HTML}{A95852}
\definecolor{AreteLime}{HTML}{6E9250}
\definecolor{AreteMist}{HTML}{F3F3F3}
\definecolor{AretePaper}{HTML}{FFFFFF}
\definecolor{AreteLine}{HTML}{C8C8C8}
\definecolor{AretePositive}{HTML}{1B1B1B}
\colorlet{AreteBlueSoft}{AreteMist}
\colorlet{AreteVioletSoft}{AreteMist}
\colorlet{AreteCyanSoft}{AreteMist}
\colorlet{AreteGoldSoft}{AreteMist}
\colorlet{AreteCoralSoft}{AreteMist}

\newcommand{\benchyes}{\textcolor{AreteBlue}{\ensuremath{\checkmark}}}
\newcommand{\benchno}{\textcolor{AreteCoral}{\ensuremath{\times}}}
\providecommand{\pfLocalNoteZH}[1]{}

\newenvironment{pfWideFigure}{%
  \begin{figure}[h]\centering
}{%
  \end{figure}
}
\newenvironment{pfWideTable}{%
  \begin{table}[h]\centering
}{%
  \end{table}
}

\renewcommand{\eqref}[1]{equation~\ref{#1}}
\newcommand{\BenchmarkName}{SWE-Shift}

\providecommand{\pfLocalNoteZH}[1]{}

\newcommand{\StudySystems}{17}
\newcommand{\EvaluatedSystems}{17}

\newcommand{\SuccessDropStatement}{Success drops in all 17 Clean--Shifted comparisons.}
\newcommand{\MissingResultsNote}{}
\newcommand{\MissingHeatmapNote}{}

\newcommand{\CleanMinimum}{10.71}
\newcommand{\CleanMaximum}{67.35}

\newcommand{\ShiftedMinimum}{1.16}
\newcommand{\ShiftedMaximum}{24.79}

\newcommand{\PSRMinimum}{0.90}
\newcommand{\PSRMaximum}{20.34}

\newcommand{\RuntimeMeasuredSystems}{17}
\newcommand{\RuntimeZeroSystems}{16}
\newcommand{\ToolMeasuredSystems}{17}
\newcommand{\ToolZeroSystems}{17}
\newcommand{\EffectsMeasuredSystems}{17}
\newcommand{\EffectsZeroSystems}{13}
\newcommand{\DisclosureMeasuredSystems}{17}
\newcommand{\DisclosureZeroSystems}{13}
\newcommand{\StaleMeasuredSystems}{17}
\newcommand{\StaleZeroSystems}{17}

\AddToHook{env/table/begin}{\setlength{\parskip}{0pt}}
\AddToHook{env/figure/begin}{\setlength{\parskip}{0pt}}
\newif\ifSWEMainResultsTable
\AddToHook{file/main-results.tex/before}{\SWEMainResultsTabletrue}
\AddToHook{file/main-results.tex/after}{\SWEMainResultsTablefalse}
\let\SWEOriginalWrapTable\wraptable
\let\SWEOriginalEndWrapTable\endwraptable
\RenewDocumentEnvironment{wraptable}{o m o m}{%
  \ifSWEMainResultsTable\setlength{\intextsep}{4pt}\fi
  \IfNoValueTF{#3}
    {\SWEOriginalWrapTable{#2}{#4}}
    {\SWEOriginalWrapTable{#2}[#3]{#4}}%
  \setlength{\parskip}{0pt}%
  \captionsetup{font=footnotesize,position=top,skip=4pt}%
  \ifSWEMainResultsTable\vspace{11pt}\fi
}{\SWEOriginalEndWrapTable}
\AddToHook{file/related-work.tex/before}{%
  \AddToHook{env/tabular/begin}[prism-glance-fit]{\setlength{\tabcolsep}{0.9pt}}}
\AddToHook{file/related-work.tex/after}{%
  \RemoveFromHook{env/tabular/begin}[prism-glance-fit]}
\AddToHook{file/related-comparison.tex/before}{%
  \AddToHook{env/tabular/begin}[prism-comparison-fit]{\fontsize{9.4}{11.28}\selectfont}}
\AddToHook{file/related-comparison.tex/after}{%
  \RemoveFromHook{env/tabular/begin}[prism-comparison-fit]}
\AddToHook{file/results.tex/before}{\FloatBarrier}
\AddToHook{file/introduction.tex/before}{%
  \AddToHookNext{env/quote/begin}{\interlinepenalty=10000}%
  \AddToHookNext{env/itemize/begin}{\csname @beginparpenalty\endcsname=10000}}
\AddToHook{file/results.tex/before}{%
  \let\SWEOriginalResultsFigure\figure
  \RenewDocumentCommand{\figure}{O{tbp}}{\SWEOriginalResultsFigure[H]}}
\AddToHook{file/results.tex/after}{\let\figure\SWEOriginalResultsFigure}

\graphicspath{{arxiv/template/}}
\title{\BenchmarkName: Evaluating Coding Agents in Evolving Workspaces}
\author[1]{Victor~Shea-Jay~Huang}
\author[5]{Rongqing~Li}
\author[2]{Zhangyue~Yin}
\author[5]{Kelong~Mao}
\author[3]{Yaqi~Zhao}
\author[2]{Xinyuan~Wang}
\author[4]{Yao~Yin}
\author[4]{Lirui~Luo}
\author[1]{Renrui~Zhang}
\author[1,\dagger]{Hongsheng~Li}
\affiliation[1]{\mbox{MMLab, The Chinese University of Hong Kong}}
\affiliation[2]{\mbox{Fudan NLP Lab, Fudan University}}
\affiliationbreak
\affiliation[3]{\mbox{The Hong Kong University of Science and Technology}}
\affiliation[4]{\mbox{Peking University}}
\affiliation[5]{\mbox{Somewhere on the Earth}}
\abstract{
Repository-level coding benchmarks largely assume that tasks and environments remain unchanged throughout execution. In software development, code, dependencies, runtime state, and requirements can change while an agent is still working. We introduce SWE-Shift, a benchmark for evaluating coding agents under such mid-execution changes. From the same saved history and workspace, a Clean continuation proceeds without intervention, while a matched Shifted continuation receives a task-relevant update. Success requires completing the current task, preserving still-valid work, and satisfying the applicable state and evidence checks. The benchmark contains 120 scenarios across 12 change families, each evaluated at nine scheduled intervention positions. We evaluate \EvaluatedSystems{} coding-agent systems, all of which have lower Shifted than Clean success. GPT-6 Astra drops from 67.35\% to 24.79\%, with 20.34\% paired success. Tool/session changes and stale evidence yield zero paired success across all systems. Late-position paired success is lower for some systems and higher for others. These results show that success on the unchanged task can coexist with failures to preserve work, recover state, or refresh evidence after an update.
}
\code{https://github.com/JeixHuang/SWE-Shift}{https://github.com/JeixHuang/SWE-Shift}
\metadata[Correspondence]{\email{jeix782@gmail.com}}
\hypersetup{pdftitle={SWE-Shift: Evaluating Coding Agents in Evolving Workspaces},
  pdfauthor={Victor Shea-Jay Huang, Rongqing Li, Zhangyue Yin, Kelong Mao, Yaqi Zhao, Xinyuan Wang, Yao Yin, Lirui Luo, Renrui Zhang, Hongsheng Li}}

\begin{document}
\begingroup
\tikzset{every picture/.style={}}
\begin{tikzpicture}[remember picture,overlay]
  \node[anchor=west,inner sep=0pt] at ([xshift=2.5cm,yshift=-1.52cm]current page.north west) {\includegraphics[width=3.5cm]{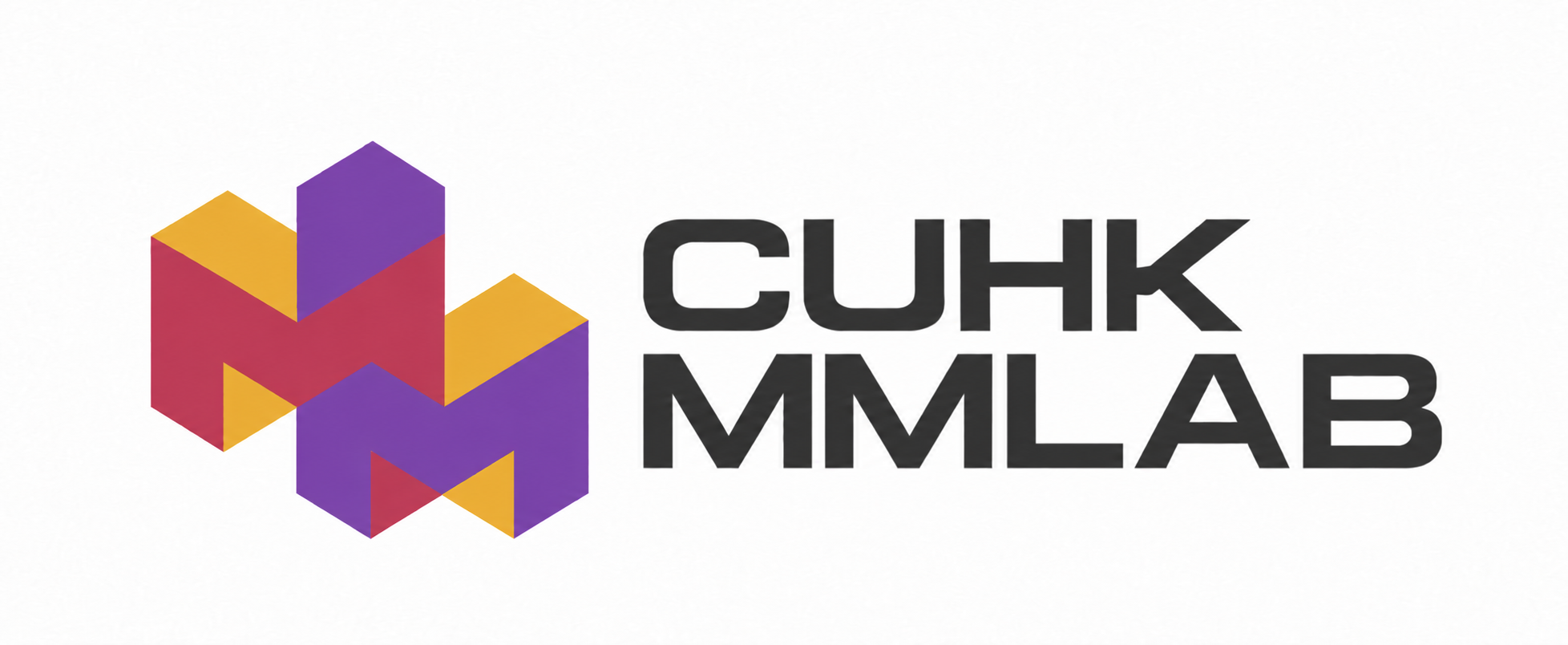}};
  \node[anchor=east,inner sep=0pt] at ([xshift=-2.5cm,yshift=-1.67cm]current page.north east) {%
    \includegraphics[width=0.56cm,height=0.56cm,keepaspectratio]{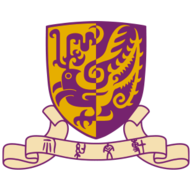}\hspace{0.20cm}%
    \includegraphics[width=0.56cm,height=0.56cm,keepaspectratio,trim=0 221.52bp 366.96bp 0,clip]{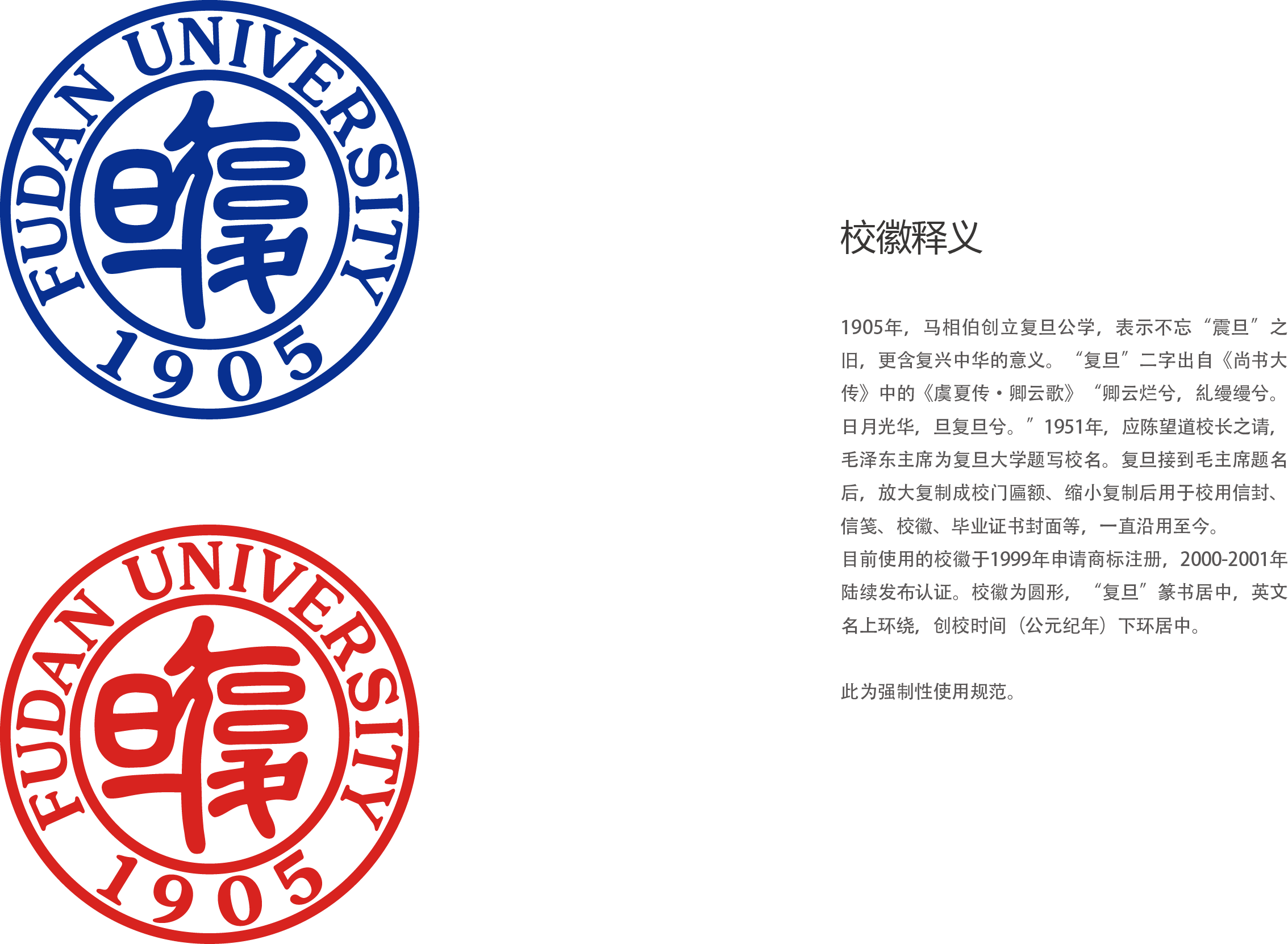}\hspace{0.20cm}%
    \includegraphics[width=0.56cm,height=0.56cm,keepaspectratio]{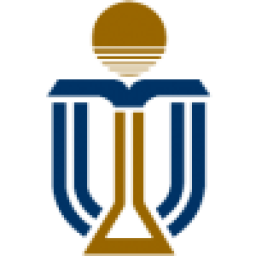}\hspace{0.20cm}%
    \includegraphics[width=0.56cm,height=0.56cm,keepaspectratio]{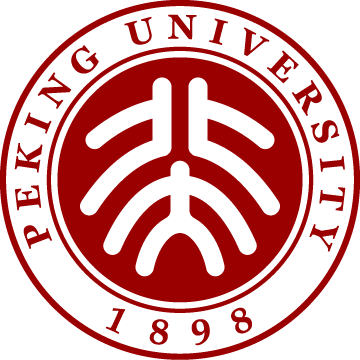}};
  \draw[gray!55,line width=0.6pt] ([xshift=2.5cm,yshift=-2.17cm]current page.north west) -- ([xshift=-2.5cm,yshift=-2.17cm]current page.north east);
\end{tikzpicture}

\vspace*{0.05cm}
\maketitle

\endgroup
\section{Introduction}
\label{sec:introduction}
\begingroup
\setlength{\parskip}{4pt}

\begingroup
\setlength{\intextsep}{5pt}
\begin{figure}[H]
\centering
\setlength{\abovecaptionskip}{2pt}
\includegraphics[width=\textwidth]{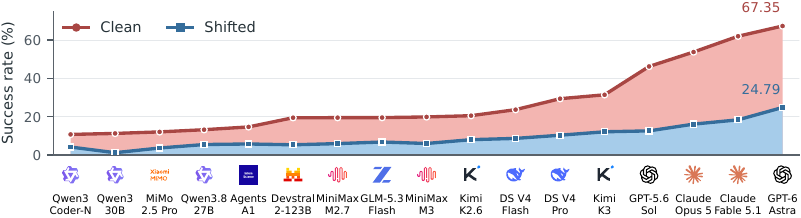}
\caption{\textbf{Success drops after workspace changes.} Clean (red) and
Shifted (blue) success rates across \EvaluatedSystems{} coding-agent systems, ordered by Clean success.}
\label{fig:main-results-teaser}
\end{figure}
\endgroup

Repository-level benchmarks such as SWE-bench~\citep{swebench} have made it possible to evaluate coding agents on realistic software engineering tasks rather than isolated code generation. An agent must understand an issue, navigate a repository, modify executable code, and satisfy tests. Yet these benchmarks largely assume that once execution begins, the task and its environment remain unchanged. Real software development is less static. Code may be edited, dependencies may change, runtime state may evolve, and requirements or available information may be updated while an agent is still working.

This raises a different reliability question:
\begin{quote}
\centering
\textbf{Can a coding agent continue to succeed when task-relevant changes occur during execution?}
\end{quote}
This question is not answered by evaluating agents on changed tasks alone. If an agent fails after a change, we do not know whether it was unable to solve the underlying task in the first place or whether the change disrupted an execution that would otherwise have succeeded. A useful evaluation must therefore separate these two sources of failure.

\begin{figure}[!h]
\centering
\setlength{\abovecaptionskip}{2pt}
\includegraphics[width=.90\textwidth]{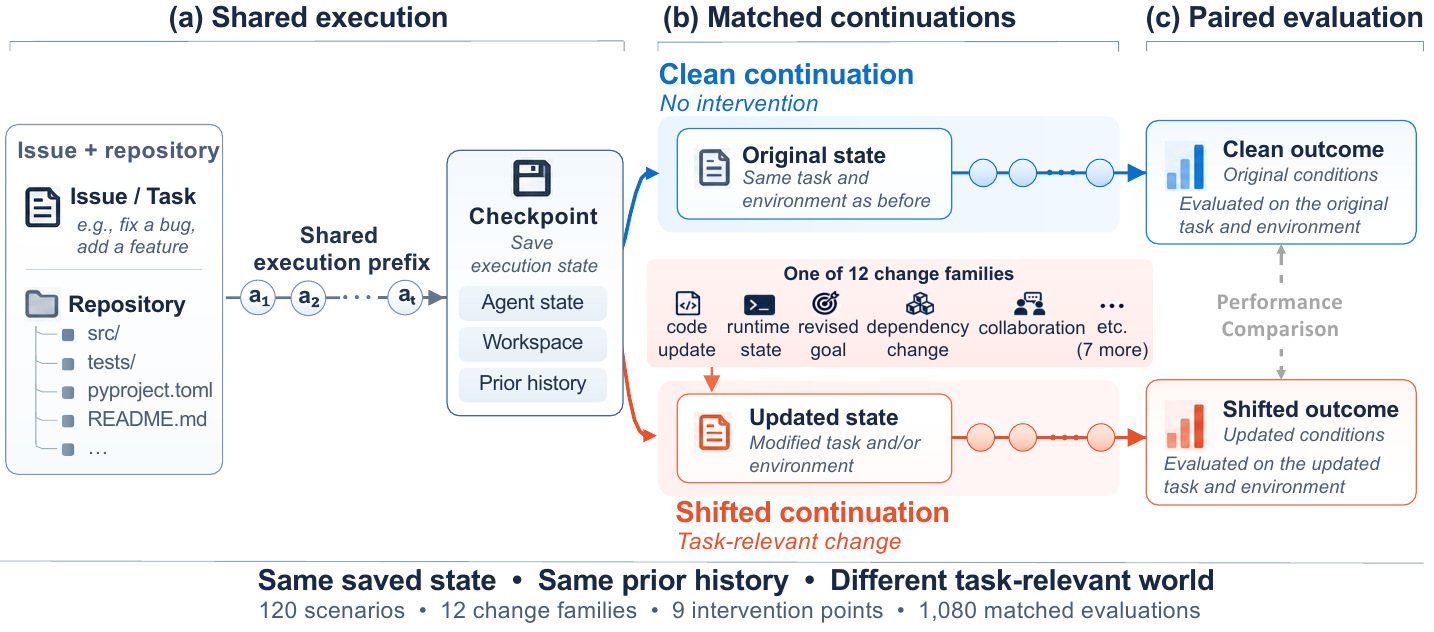}
\caption{\textbf{Per-system evaluation pipeline of SWE-Shift.} From the same saved execution state, the agent resumes in two matched continuations: Clean proceeds without intervention, while Shifted introduces a task-relevant change. Outcomes are evaluated under the original and updated conditions, respectively.}
\label{fig:intro-pipeline}
\vspace{-10pt}
\end{figure}

We introduce SWE-Shift, which addresses this problem through matched Clean and Shifted continuations (Figure~\ref{fig:intro-pipeline}). The agent first works on the task normally. At an intermediate point, we save its execution state and continue from the same prefix in two ways: the Clean continuation proceeds without a change, while the Shifted continuation introduces a task-relevant change. Because the two continuations share the same prior actions, workspace, and observable history, Clean provides a matched baseline for whether the agent can succeed from that state, while Shifted tests whether that success survives the change.

Constructing such an evaluation also requires the changes themselves to be meaningful. A change should matter to the task, but it should not make the task impossible. SWE-Shift therefore includes scenarios in which ignoring the change can invalidate an otherwise successful continuation, while a correct continuation remains possible after the change. We also introduce each change at multiple points during execution, rather than tying the benchmark to a single moment in the agent's trajectory. Finally, success is judged against the requirements that apply after the update, including preservation and evidence requirements that the original task tests may not cover.

SWE-Shift contains 120 scenarios across 12 change families, covering code and dependency updates, collaboration, runtime and session state, external effects, revised goals, and stale or misbound evidence. Nine intervention positions per scenario yield 1,080 scheduled evaluation cells per system.

Our evaluation of current coding agents reveals a substantial gap between success in unchanged and changing environments. As shown in Figure~\ref{fig:main-results-teaser}, all \EvaluatedSystems{} evaluated coding-agent systems have lower observed success after the intervention. GPT-6 Astra, which has the highest Clean success, drops from 67.35\% on Clean continuations to 24.79\% on Shifted continuations. The size of the gap also varies across change families and intervention positions, suggesting that robustness to mid-execution change captures behavior that static task success alone does not reveal.

Our contributions are summarized as follows:

\begingroup
\setlength{\parskip}{0pt}
\setlength{\topsep}{4pt}
\setlength{\partopsep}{0pt}
\setlength{\itemsep}{3pt}
\setlength{\parsep}{0pt}
\begin{itemize}
\item We identify \textbf{robustness to mid-execution change} as an evaluation target beyond static task success, distinguishing failures on the underlying task from losses of successful continuations after a task-relevant change.

\item We introduce SWE-Shift, a matched-continuation benchmark with 120 scenarios across 12 change families and nine intervention positions, enabling controlled comparison of Clean and Shifted executions from the same intermediate agent state.

\item We evaluate \EvaluatedSystems{} coding-agent systems and find a substantial gap between Clean and Shifted success, with failures varying across change families and intervention positions. Recorded trajectories and diagnostic controls further characterize these failures.
\end{itemize}
\endgroup
\endgroup

\Needspace{16\baselineskip}
\section{Related Work}
\label{sec:related}

\begingroup
\setlength{\columnsep}{7pt}
\setlength{\intextsep}{5pt}
\begin{wraptable}[10]{r}{.58\textwidth}
\vspace{-6pt}
\centering
\caption{\textbf{Benchmark design at a glance.}}
\label{tab:benchmark-at-glance}
\begingroup\scriptsize
\setlength{\tabcolsep}{1.3pt}
\renewcommand{\arraystretch}{1.08}
\begin{tabular}{@{}lccccccc@{}}
\toprule
Benchmark & \makecell{Mid-\\episode} & \makecell{Same\\prefix} & \makecell{Cross-\\plane} & \makecell{Preserve/\\harm} & \makecell{Multi-\\position} & \makecell{Trajectory\\oracle} & \makecell{Fresh\\evidence} \\
\midrule
SpecPath & \benchno & \benchno & \benchno & \benchno & \benchno & \benchno & \benchno \\
SyncMind & \benchno & \benchno & \benchno & \benchno & \benchno & \benchno & \benchno \\
SWE-INTERACT & \benchyes & \benchno & \benchno & \benchno & \benchno & \benchno & \benchno \\
ToolSandbox & \benchno & \benchno & \benchno & \benchyes & \benchno & \benchyes & \benchno \\
SWE-Touch & \benchyes & \benchno & \benchno & \benchyes & \benchyes & \benchno & \benchno \\
Gaia2 & \benchyes & \benchno & \benchyes & \benchyes & \benchno & \benchyes & \benchno \\
\textbf{SWE-Shift (Ours)} & \benchyes & \benchyes & \benchyes & \benchyes & \benchyes & \benchyes & \benchyes \\
\bottomrule
\end{tabular}
\endgroup
\vspace{-6pt}
\end{wraptable}

Repository benchmarks such as SWE-bench evaluate issue resolution but normally
keep the task and environment fixed during execution~\citep{swebench}.
Related dynamic coding evaluations study different kinds of change: SWE-Touch injects validated user edits during execution,
SyncMind begins recovery from repository inconsistencies, SWE-INTERACT reveals
or revises requirements through dialogue, and SpecPath varies specification
history before implementation~\citep{swetouch,syncmind,sweinteract,specpath}.
ToolSandbox contributes state dependencies, milestone/Minefield trajectory
evaluation, while Gaia2 adds independently evolving asynchronous events and
action-level write verification~\citep{toolsandbox,gaia2}. \Cref{tab:benchmark-at-glance} compares these evaluation designs. ``Mid-episode'' denotes an external intervention after execution begins. ``Fresh evidence'' means that completion claims are checked against
the current state. SWE-Shift combines three controls: matched continuations from the same
recorded prefix, changes to multiple parts of the task environment, and checks
that reject the loss of still-valid behavior even when the original task tests
pass. \Cref{tab:related-comparison} in Appendix~\ref{app:related} provides a detailed comparison.
\par
\endgroup

\section{SWE-Shift Benchmark}
\label{sec:benchmark}

We first describe how task-relevant changes are designed (Section~\ref{sec:change-design}).
We then explain how matched continuations and intervention timing support controlled comparisons (Section~\ref{sec:matched-continuation}).
Finally, we define the success criteria and metrics used to evaluate these continuations (Section~\ref{sec:success-metrics}).

\subsection{Designing Task-Relevant Changes}
\label{sec:change-design}

SWE-Shift builds 120 change scenarios on 23 task-root snapshots from 17 repositories,
drawn from SWE-bench Verified~\citep{swebench_verified},
Multi-SWE-bench~\citep{multiswebench}, and
SWE-bench Multilingual~\citep{swebench_multilingual}.
Each scenario combines an existing repository task with a task-relevant change introduced during execution.
\Cref{tab:task-sources} in Appendix~\ref{sec:task-sources} provides the scenario counts by source.

To evaluate adaptation, these changes must challenge the agent without making successful completion impossible.
We therefore require each change to be both \emph{relevant} and \emph{recoverable}.
Relevance means that ignoring the change can invalidate an otherwise successful continuation.
Recoverability means that, after the change, at least one valid way to complete the task remains available through the agent-accessible tools.
Together, these requirements exclude changes that do not matter to the task and changes that make the task unsolvable.

To turn such a change into an executable evaluation scenario, we specify four elements:
the initial repository task, an authorized mid-execution change, a way for the agent to encounter or discover that change, and executable checks for a correct continuation.
These checks reflect the requirements that apply after the update: some original requirements remain in force, while others may be added or withdrawn.
A correct continuation must satisfy the current requirements while preserving work and effects that remain valid.

Using this scenario structure, SWE-Shift covers 12 change families, with ten scenarios in each family.
Figure~\ref{fig:change-families} organizes them into three broad groups according to what the change affects.
\begin{pfWideFigure}
\centering
\includegraphics[width=\linewidth]{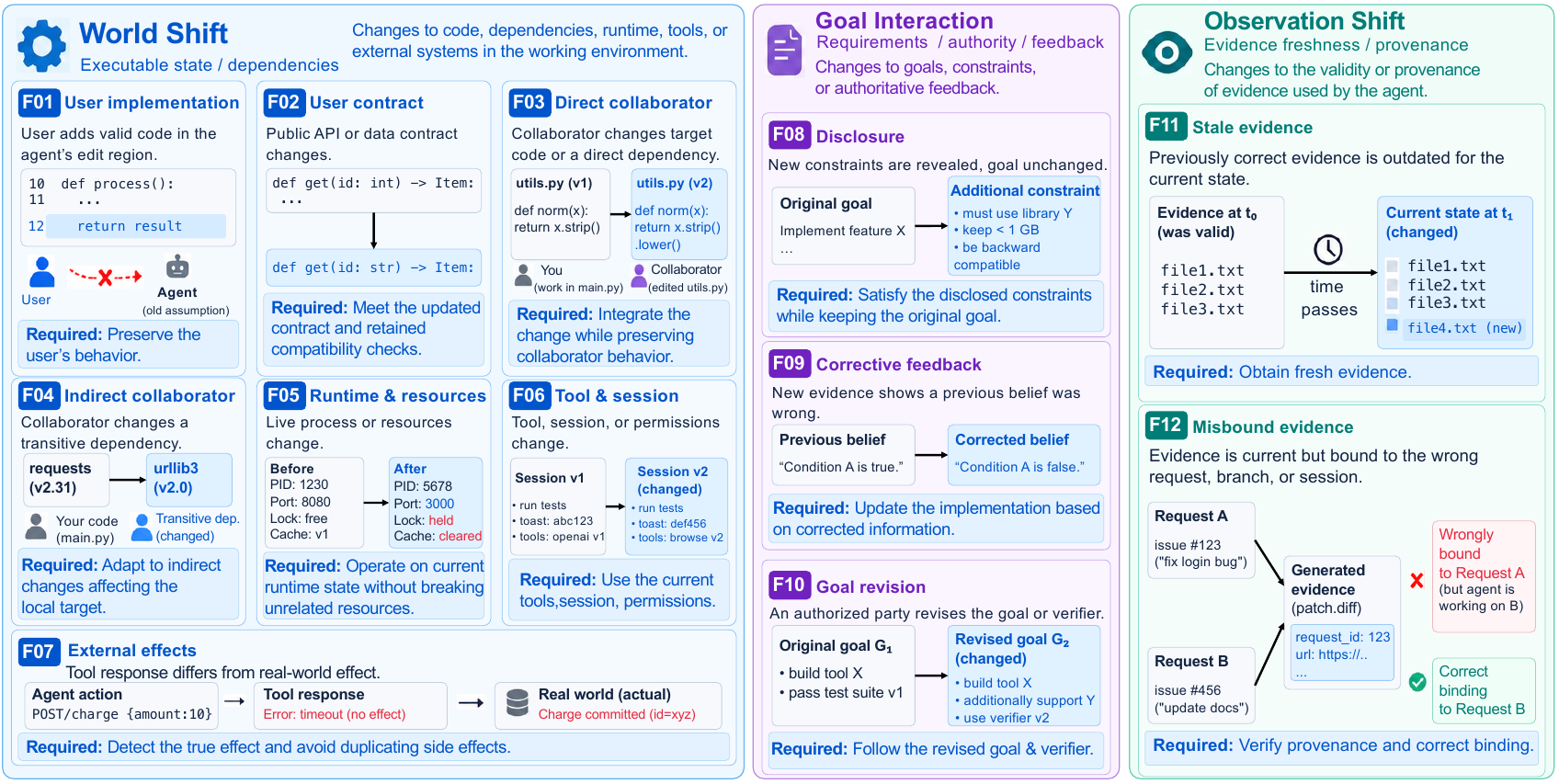}
\caption{\textbf{What SWE-Shift evaluates.}
The 12 change families are organized into three broad groups. World Shift
(F01--F07) tests changes to executable state or dependencies. Goal Interaction
(F08--F10) tests newly disclosed requirements, corrective feedback, and revised
goals. Observation Shift (F11--F12) tests stale or misbound evidence. Each
panel illustrates the intervention and the required continuation.}
\label{fig:mechanisms}
\label{fig:change-families}
\end{pfWideFigure}

F01--F07 modify the executable world or its dependencies, including user and collaborator edits, interface changes, runtime and session state, and external effects.
F08--F10 change what the agent is required to act on through newly disclosed constraints, corrective feedback, or an authorized revision of the goal.
F11--F12 change the validity or binding of evidence: an observation may describe an earlier state or refer to the wrong request, branch, or session.

\subsection{Matched Continuation Protocol}
\label{sec:matched-continuation}

SWE-Shift compares continuations with and without a mid-execution change from the same prior execution (\Cref{fig:intro-pipeline}).
For each system, scenario, and intervention position, the agent first executes a single shared prefix.
At the scheduled checkpoint, the runner saves the agent-visible history, workspace, controller state, available tools, system settings, and execution counters.
The saved state is then restored twice to create a matched pair of continuations.

In the \textbf{Clean} continuation, execution proceeds without the scenario's task-relevant update.
In the \textbf{Shifted} continuation, the authorized update is applied at the fork.
Up to that point, the two continuations have the same recorded execution history and workspace state. Only after the intervention can their observations and actions diverge.
This differs from comparing independently generated runs, where differences before the change may already affect the outcome.
The Clean continuation therefore provides a matched baseline for whether the agent can complete the task from that particular intermediate state, while the Shifted continuation tests whether that success survives the change.

The update is introduced through the scenario's declared interface.
Depending on the case, the agent may encounter it as a workspace change, a delivered message, a changed tool or session response, or new information returned by an allowed state query.
The evaluator may use additional held-out checks to determine correctness, but these checks do not provide additional information to the agent.
Thus, the paired comparison changes the task-relevant conditions after the fork while keeping the observable pre-intervention execution fixed.

\subsubsection{Intervention Timing}
\label{sec:intervention-timing}

A single intervention point could make the evaluation depend strongly on when the change happens to occur.
SWE-Shift therefore evaluates every scenario at nine progressively later intervention positions, denoted T10--T90.
These positions are constructed from an eligible prefix of a reference solution, ending before the first action whose correctness depends on the update.
The construction procedure provides nine distinct intervention anchors spanning 10\%--90\% of the eligible reference progress for every admitted scenario.

During model evaluation, the intervention position determines how many agent actions are executed before the fork. The evaluated agent is not required to follow the reference trajectory used to construct the schedule.
Consequently, T10--T90 represent progressively later \emph{scheduled positions}, rather than normalized semantic progress through the task.
Different systems may therefore have reached different workspace states when the same nominal intervention position is triggered.
If an agent terminates before the scheduled intervention or changes the state in a way that affects whether the update can still be applied, the cell retains its assigned intervention position and is handled by the benchmark's outcome protocol without rescheduling.

\subsection{Success Criteria and Metrics}
\label{sec:success-metrics}

A continuation is counted as successful only if it satisfies every check that applies under the conditions in which it ends.
For the Clean continuation, these are the original task and environment conditions. For the Shifted continuation, they are the conditions after the update.
The original task tests are therefore only one part of the evaluation.
Depending on the scenario, success may additionally require preserving still-valid behavior, following the current goal or interface, avoiding prohibited effects, or obtaining evidence that is valid for the current state.
The implementation groups these requirements into task, preservation, recovery, authority, effect, evidence, budget, and terminal-integrity checks.
The criterion is conjunctive: failing any applicable check makes the continuation unsuccessful. \Cref{tab:checker-map} summarizes the required evidence.

For model--agent system $m$, scenario $i$, and intervention position $t$, let
$C_{mit}\in\{0,1\}$ and $S_{mit}\in\{0,1\}$ denote verified success in the Clean and Shifted continuations, respectively.
Ordinary paired credit requires success in both continuations:
\begin{equation}
R_{mit} = C_{mit} S_{mit}.
\label{eq:paired-success}
\end{equation}
This joint-success component measures success in both observed continuations: under the original conditions and after the task-relevant change.

Some agents complete the required task before the scheduled intervention is reached.
Let $P_{mit}\in\{0,1\}$ denote verified pre-intervention completion under the benchmark's early-completion procedure.
The score assigned to an intervention cell is
\begin{equation}
K_{mit} = \max\{R_{mit}, P_{mit}\}.
\label{eq:cell-score}
\end{equation}
Thus, a cell receives credit either because both matched continuations succeed or because the task was already completed before the agent could be exposed to the change.

We aggregate scenarios hierarchically: each of the 12 change families receives equal weight, followed by equal weight for clusters within a family and scenarios within a cluster.
Let $w_i$ denote the resulting normalized scenario weight, and let
$\mathcal{T}=\{\mathrm{T10},\ldots,\mathrm{T90}\}$ denote the nine intervention positions.
We report
\par
\begingroup
\setlength{\abovedisplayskip}{3pt}
\setlength{\belowdisplayskip}{3pt}
\setlength{\abovedisplayshortskip}{3pt}
\setlength{\belowdisplayshortskip}{3pt}
\thinmuskip=2mu
\thickmuskip=3mu
\noindent
\begin{minipage}[t]{.32\linewidth}
\begin{equation}
\mathrm{Clean}_m=\frac{1}{9}\sum_{\substack{t\in\mathcal{T}\\i}}w_i C_{mit},
\end{equation}
\end{minipage}\hfill
\begin{minipage}[t]{.345\linewidth}
\begin{equation}
\mathrm{Shifted}_m=\frac{1}{9}\sum_{\substack{t\in\mathcal{T}\\i}}w_i S_{mit},
\end{equation}
\end{minipage}\hfill
\begin{minipage}[t]{.315\linewidth}
\begin{equation}
\mathrm{PSR}_m=\frac{1}{9}\sum_{\substack{t\in\mathcal{T}\\i}}w_i K_{mit}.
\end{equation}
\end{minipage}\par
\endgroup

Our primary paired metric is the \textbf{Paired Success Rate (PSR)}.
PSR averages credit for joint success or verified completion before intervention.

We additionally report the \textbf{Conditional Retention Rate (CRR)}, which asks what fraction of Clean success is also retained in the Shifted continuation:
\begin{equation}
\mathrm{CRR}_m
=
\frac{
\sum_{t\in\mathcal{T}}\sum_i w_i R_{mit}
}{
\sum_{t\in\mathcal{T}}\sum_i w_i C_{mit}
},
\label{eq:crr}
\end{equation}
when the denominator is nonzero.
Its numerator is joint success $R$ and excludes the early-completion credit in PSR.
Clean, Shifted, PSR, and CRR therefore summarize baseline continuation success, success under changed conditions, paired success, and retention, respectively.

\section{Experiments}
\label{sec:setup}
\label{sec:results}

Our experiments ask whether coding capability survives a workspace change
(RQ1), which changes expose shared or system-specific weaknesses (RQ2), whether
their timing matters (RQ3), and what a failed continuation can hide from
ordinary task tests (RQ4). We then examine how state handling and post-fork token
budgets affect success.

\subsection{Experimental setup}

We evaluate \StudySystems{} coding-agent systems, listed in
\Cref{tab:system-config}, using the same 120 scenarios and nine intervention positions,
with the same available tools
and post-fork token budget $B=1{,}000{,}000$ tokens per continuation, counted as cumulative
input plus output tokens across all post-fork model calls. Each row in
\Cref{tab:main-results} denotes a deployed model--agent system.
Provider-hosted systems use their native agent
applications: Codex app for GPT-5.6 Sol and GPT-6 Astra, and Claude
Code app for Claude Opus 5 and Claude Fable 5.1, all at
\texttt{xhigh} reasoning. The
open-weight coding-agent systems were served locally on four compute nodes with
eight NVIDIA B300 GPUs each (32 GPUs total).

Clean, Shifted, PSR, and CRR use the family--cluster--scenario weighting in
\Cref{sec:benchmark}. The reported main results are point values.


\Needspace{27\baselineskip}
\subsection{RQ1: Does coding capability survive a workspace change?}
\label{sec:rq-capability}

\begin{wraptable}[16]{r}{.63\textwidth}
\vspace{-23pt}
\centering
\caption{\textbf{Continuation performance (\%).} \MissingResultsNote{}
Full identities appear in \Cref{tab:system-config}.}
\label{tab:main-results}
\begingroup\scriptsize\setlength{\tabcolsep}{1.4pt}\renewcommand{\arraystretch}{.96}
\begin{tabular*}{\linewidth}{@{}l@{\extracolsep{\fill}}rrrr@{}}
\toprule
System & Cln. $\uparrow$ & Shf. $\uparrow$ & PSR $\uparrow$ & CRR $\uparrow$ \\
\midrule
GPT-6 Astra~\citeyearpar{card_gpt6astra} & 67.35 & 24.79 & 20.34 & 30.20 \\
Claude Fable 5.1~\citeyearpar{card_claude_fable51} & 62.01 & 18.33 & 16.88 & 26.02 \\
Claude Opus 5~\citeyearpar{card_claude_opus5} & 53.76 & 16.09 & 14.78 & 26.54 \\
GPT-5.6 Sol~\citeyearpar{card_gpt56} & 46.22 & 12.56 & 11.30 & 23.94 \\
Kimi-K3~\citeyearpar{card_kimi3} & 31.41 & 12.09 & 8.20 & 25.12 \\
DeepSeek V4-Pro~\citeyearpar{card_deepseekv4pro} & 29.37 & 10.29 & 8.83 & 29.15 \\
DeepSeek V4-Flash~\citeyearpar{card_deepseekv4flash} & 23.65 & 8.63 & 7.60 & 30.99 \\
Kimi-K2.6~\citeyearpar{card_kimi26} & 20.50 & 7.97 & 7.10 & 34.63 \\
MiniMax-M3~\citeyearpar{card_minimaxm3} & 19.84 & 5.97 & 6.20 & 29.08 \\
GLM-5.3-Flash~\citeyearpar{card_glm53flash} & 19.50 & 6.75 & 5.30 & 25.18 \\
MiniMax-M2.7~\citeyearpar{card_minimax27} & 19.47 & 5.91 & 5.70 & 27.84 \\
Devstral 2-123B~\citeyearpar{card_devstral2} & 19.44 & 5.29 & 4.70 & 22.07 \\
Agents-A1~\citeyearpar{card_agentsa1} & 14.63 & 5.72 & 4.50 & 29.01 \\
Qwen3.8-27B~\citeyearpar{card_qwen3827b} & 13.17 & 5.43 & 4.34 & 30.45 \\
MiMo V2.5-Pro~\citeyearpar{card_mimo25pro} & 12.03 & 3.61 & 2.60 & 21.26 \\
Qwen3-30B-A3B~\citeyearpar{card_qwen30} & 11.23 & 1.16 & 0.90 & 6.02 \\
Qwen3-Coder-Next~\citeyearpar{card_qwencoder} & 10.71 & 4.12 & 2.10 & 14.97 \\
\bottomrule

\end{tabular*}\endgroup
\vspace{-7pt}
\end{wraptable}

\paragraph{Workspace changes reduce continuation success.}
\SuccessDropStatement{}
Across the \EvaluatedSystems{} coding-agent systems in \Cref{tab:main-results},
Clean success ranges from \CleanMinimum\% to \CleanMaximum\%, compared with
\ShiftedMinimum--\ShiftedMaximum\% for Shifted and
\PSRMinimum--\PSRMaximum\% for PSR.
GPT-6 Astra attains 67.35\% Clean, 24.79\% Shifted, and 20.34\% PSR.
Its CRR of 30.20\% means that fewer than one-third of its weighted Clean
successes also succeed under change.

\paragraph{Higher Clean success does not imply higher retention.}
In \Cref{tab:main-results}, Kimi-K3 attains 31.41\% Clean success, 8.20\% PSR, and 25.12\% CRR.
DeepSeek-V4-Pro has lower Clean success (29.37\%) yet higher PSR (8.83\%)
and CRR (29.15\%). Kimi-K2.6 reaches 34.63\% CRR despite a
lower Clean score of 20.50\%. CRR measures retention conditional on Clean success and complements the
absolute continuation scores.

\subsection{RQ2: Which changes are persistent failure points?}
\label{sec:rq-family}

\paragraph{Tool/session changes and stale evidence expose shared weaknesses.}
Under the common evaluation setting, \Cref{fig:when-what}(a) shows the
following family-level PSR results:
\begin{enumerate}
  \item \textbf{Tool and session changes (F06).} A tool, authorization, schema,
  or session protocol changes after execution begins, invalidating earlier
  calls. The agent must discover and use the current permitted interface while
  preserving unrelated session state. PSR is 0.00\% for \ToolZeroSystems{} of
  \ToolMeasuredSystems{} systems.
  \item \textbf{Stale evidence (F11).} Evidence refers to an outdated state.
  The agent must obtain evidence bound to the current state
  before editing or accepting completion.
  PSR is 0.00\% for \StaleZeroSystems{} of \StaleMeasuredSystems{} systems.
  \item \textbf{Runtime and resource changes (F05).} A process, lock, port,
  cache, or resource owner changes while the task is in progress. The agent
  must rebind to the live state and recover without harming unrelated
  resources. MiMo-V2.5-Pro obtains 0.60\% PSR.
  \RuntimeZeroSystems{} of \RuntimeMeasuredSystems{} systems score zero.
  \item \textbf{Disclosure changes (F08).} A previously hidden constraint or
  fact becomes visible during execution. \DisclosureZeroSystems{} of
  \DisclosureMeasuredSystems{} systems score 0.00\%
  PSR. Claude Fable 5.1 reaches 16.20\%, while GPT-6 Astra reaches 14.90\%.
\end{enumerate}

\paragraph{The remaining families reveal selective strengths.}
In \Cref{fig:when-what}(a), Kimi-K3 reaches 63.50\% on F01 (User implementation), while Claude Fable 5.1
reaches 60.80\%. On F04 (Indirect collaborator), GPT-6 Astra scores 57.30\%
and GPT-5.6 Sol scores 39.30\%. On F07 (External effects), GPT-6 Astra reaches
22.10\% and GLM-5.3-Flash reaches 18.50\%. \EffectsZeroSystems{} of
\EffectsMeasuredSystems{} systems score zero.
A higher overall PSR does not imply higher PSR in every change family.

\Needspace{9\baselineskip}
\subsection{RQ3: Does the timing of a change affect continuation?}
\label{sec:rq-timing}

\paragraph{Late-position PSR can be lower or higher.}
\Cref{fig:when-what}(b) shows PSR at the nine scheduled intervention positions.
Comparing mean PSR at T70--T90 with T10--T30, GPT-6 Astra loses 20.05
percentage points, Claude Fable 5.1 loses 15.43, GPT-5.6 Sol loses 12.53, and Claude Opus 5 loses 11.93
points. GLM-5.3-Flash
instead gains 2.97 points, Kimi-K2.6 gains 2.20 points, and
DeepSeek-V4-Flash gains 1.60 points.

\paragraph{Timing profiles differ across systems.}
The timing profiles in \Cref{fig:when-what}(b) show that mean PSR at T70--T90 is lower than at T10--T30 for both GPT systems,
both Claude systems, and both MiniMax systems. The corresponding mean is
higher for GLM-5.3-Flash, both DeepSeek systems, and both Kimi systems.
These comparisons summarize scheduled positions within each system's trajectory.

\begin{pfWideFigure}
\centering
\includegraphics[width=\linewidth]{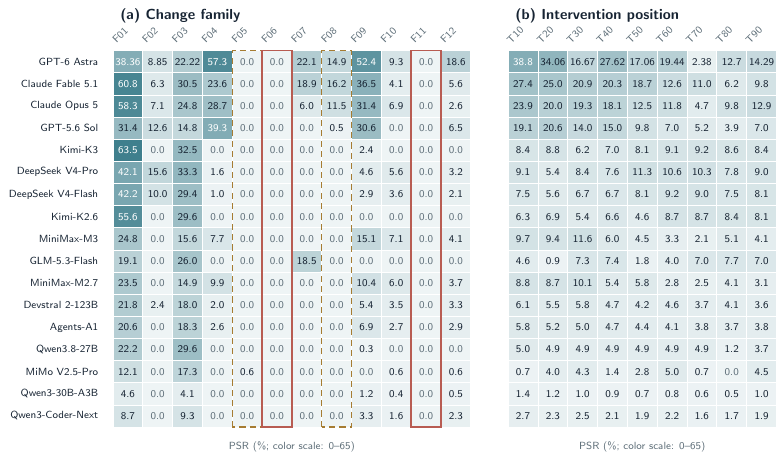}
\caption{\textbf{Paired success across changes and intervention positions.} PSR by change family (a) and
intervention position (b). \MissingHeatmapNote{}
Solid coral outlines highlight tool/session changes (F06) and stale evidence (F11).
Dashed gold outlines highlight runtime/resource changes (F05) and disclosure
(F08).}
\label{fig:timing}
\label{fig:families}
\label{fig:when-what}
\end{pfWideFigure}


\subsection{RQ4: What do task tests miss after an update?}
\label{sec:rq-obligations}

After an update, passing the original tests or completing one corrective
action may still leave other requirements unmet. We examine three cases from
the 24-case qualitative audit set (Appendix~\ref{sec:atlas}), using the recorded
GPT-5.6 Sol actions and evaluator results to explain the final outcomes.

\paragraph{Tests can pass after a collaborator's change is removed.}
In the Django case, a collaborator adds an API that controls when migration
operations can be combined. The agent initially uses this API, then removes
it while fixing the original issue. All 39 original tests pass, but the
preservation check reports that the collaborator's API is missing.

\paragraph{Compensation can succeed while workspace checks fail.}
In the Astropy case, a cancellation arrives after an external effect has
already occurred. The agent compensates for that effect, and the tool record
confirms that compensation succeeded. The workspace-preservation check still fails.

\paragraph{The final edits need fresh evidence.}
In the scikit-learn case, the agent restores access to current task evidence,
then edits the code without querying that source again. The original task
checks pass, but the required fresh observation is missing. The patch also
exceeds the allowed edit scope.

SWE-Shift therefore evaluates both the final workspace and the recorded
actions against the requirements that apply after the update.

\subsection{Diagnostic controls}
\label{sec:ablation}

\begin{figure}[!t]
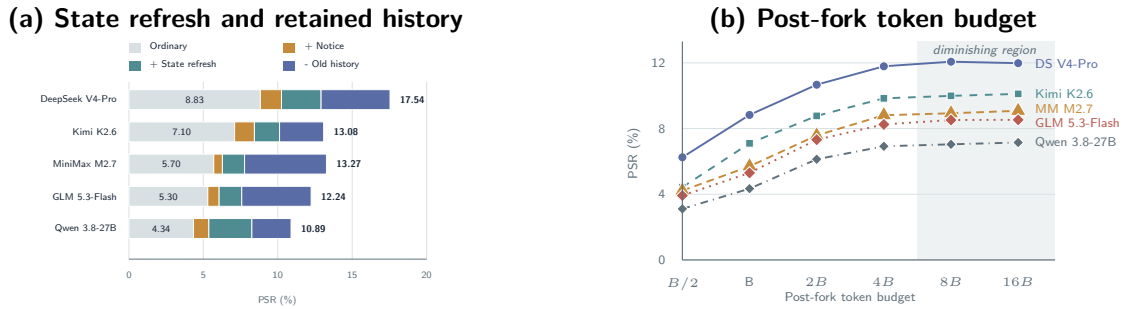
\centering
\input{content/state-refresh}\hfill
\input{figures/recovery-budget}
\caption{\textbf{Effects of state handling and post-fork token budget on PSR.} (a) Bars show
cumulative state-access interventions from the Table~\ref{tab:main-results}
baseline to the final PSR. (b) Curves show PSR across post-fork token budgets.
$B$ is the baseline of 1{,}000{,}000 input plus output tokens per continuation, and the
gray region marks $8B$--$16B$.}
\label{fig:diagnostic-controls}
\label{fig:state-refresh-ablation}
\label{fig:budget-ablation}
\end{figure}

\Cref{fig:diagnostic-controls} shows how state access, retained history, and
post-fork token budget affect PSR in five systems, using the corresponding
PSR values in \Cref{tab:main-results} as baselines.

\subsubsection{State refresh and retained history}
\label{sec:state-ablation}

{\raggedright
\textbf{Adding history reset yields the largest incremental gain in four of five systems.}
In the state-access control (\Cref{fig:diagnostic-controls}(a)), we hold tasks, obligations, workspaces, tools,
and the post-fork token budget fixed. Starting from the PSR baseline in
\Cref{tab:main-results} (Ordinary), we cumulatively add a change notice,
a read-only state refresh, and removal of pre-change dialogue. Adding history reset contributes
5.50 points for MiniMax-M2.7. Qwen3.8-27B gains more from state
refresh (2.89 points) than from history reset (2.63).
The complete intervention adds 5.98--8.71 PSR points, but final PSR remains
10.89--17.54\%.
The increments measure each addition to the preceding intervention setting.
\par}

\subsubsection{Post-fork token budget}
\label{sec:budget-ablation}

{\raggedright
\textbf{Gains diminish beyond $4B$.}
In a separate control (\Cref{fig:diagnostic-controls}(b)), we test sensitivity to the post-fork token budget,
holding tasks, updates, prefixes, tools, decoding, and scorer fixed. We scale
the budget for each continuation from $B/2$ to $16B$, where
$B=1{,}000{,}000$ cumulative input plus output tokens. This tests
500{,}000--16{,}000{,}000 tokens per continuation.
The curves report absolute PSR across budgets. The $B$
point is the same baseline reported in \Cref{tab:main-results}.
Each doubling up to $4B$ adds 0.79--2.67 points across the five systems.
By contrast, $4B\!\to\!8B$ adds 0.12--0.28 points, and
$8B\!\to\!16B$ changes PSR by $-0.09$ to $+0.15$ points.
Thus quadrupling the post-fork token budget beyond $4B$ adds only 0.19--0.28 points in total,
while PSR at $16B$ remains 7.15--11.98\%.
Increasing the post-fork token budget helps at smaller values, but yields
little further improvement in PSR within the tested range.
\par}

\section{Implications and Limitations}
\label{sec:implications}
\label{sec:limitations}

The family results in \Cref{fig:when-what}(a) identify tool/session changes and stale evidence as shared
failure points. The trajectory analysis in \Cref{sec:rq-obligations} shows how agents can pass original
task tests while removing valid collaborator work or relying on outdated
evidence. These failures motivate checking preservation and evidence validity
alongside task completion.

The evaluation covers 120 designed scenarios, some sharing task roots, with
one update per run. Diagnostic controls cover five systems. Pairing fixes the
recorded history and workspace state, while model randomness remains
uncontrolled. Intervention position varies with execution stage, accumulated
context, and remaining work. The post-fork token budget is fixed. Freshness checks verify
required observations. Task and preservation checks assess the resulting
state (Appendix~\ref{app:measurement}). Evaluating repeated changes in natural
software-development workflows is a next step.

\section{Conclusion}

Existing repository-level coding benchmarks largely assume fixed tasks and
environments, leaving agents' ability to handle mid-execution changes
insufficiently tested. We introduced SWE-Shift to address this gap through
task-relevant, recoverable interventions and matched Clean and Shifted
continuations from the same saved execution state. Its success criteria jointly
assess task completion, preservation of valid work, recovery, and evidence
validity. Evaluation across \EvaluatedSystems{} coding-agent systems shows that
success under unchanged conditions often does not survive an update.
Family-level and trajectory analyses further reveal recurring failures in state
recovery, preservation, and evidence use that original task tests can miss.
SWE-Shift provides a common testbed for measuring robustness to change,
diagnosing these failures, and assessing whether new methods improve agents'
ability to continue working reliably in evolving software environments.

\FloatBarrier
\subsection*{AI use statement}

OpenAI Codex assisted with manuscript organization, language editing, LaTeX
formatting, writing code for figures and tables, reference verification, and
benchmark data selection and construction.

For the benchmark experiments, we used Codex app for GPT-5.6 Sol and GPT-6
Astra, and Claude Code app for Claude Opus 5 and Claude Fable 5.1, all at
\texttt{xhigh} reasoning. The open-weight coding-agent systems were served
locally on four compute nodes with eight NVIDIA B300 GPUs each (32 GPUs total).

\bibliographystyle{iclr2027_conference}
\bibliography{references}

\appendix

\clearpage
\section*{Appendix Contents}
\phantomsection
\pdfbookmark[0]{Appendix Contents}{appendix-contents}
\setcounter{tocdepth}{2}
\makeatletter
\@starttoc{atoc}
\makeatother
\clearpage

\makeatletter
\newcommand{\pfAppendixAddContentsLine}[3]{%
  \def\pfAppendixEntryLevel{#2}%
  \def\pfAppendixSectionLevel{section}%
  \def\pfAppendixSubsectionLevel{subsection}%
  \ifx\pfAppendixEntryLevel\pfAppendixSectionLevel
    \addtocontents{atoc}{%
      \protect\contentsline{#2}{#3}{\thepage}{\@currentHref}%
    }%
  \else\ifx\pfAppendixEntryLevel\pfAppendixSubsectionLevel
    \addtocontents{atoc}{%
      \protect\contentsline{#2}{#3}{\thepage}{\@currentHref}%
    }%
  \fi\fi
}
\makeatother
\begingroup
\let\pfSavedAddContentsLine\addcontentsline
\let\addcontentsline\pfAppendixAddContentsLine
\section{Extended Method}
\label{app:method-details}

\subsection{What changes, and what must survive?}

A scenario specifies a repository task, a task-relevant change, how the agent
can discover that change, and checks for a correct continuation. The change
may affect repository files, runtime or session state, external effects,
requirements, or the evidence available to the agent. After the change, some
requirements remain, some are added, and some may be withdrawn. A correct
continuation must satisfy the current requirements while preserving work and
effects that are still valid.

The suite covers ten scenarios in each of 12 change families
(\Cref{fig:mechanisms}). F01--F07 concern code, collaborators, runtime,
sessions, and effects; F08--F10 separate disclosure, corrective feedback, and
goal revision; F11--F12 separate stale from misbound evidence. The
120 scenarios comprise an initial 24-case set and a 96-case expansion, sharing
23 root snapshots and 39 family--archetype--root clusters. The initial 24 cases
were constructed and audited with GPT-5.6 Sol in the Codex app at \texttt{xhigh} reasoning;
their full trajectories support the qualitative analysis in
Appendix~\ref{sec:atlas}. Roots can occur across families, and equal family
weights provide controlled coverage of the change families.

\subsection{What is held fixed at the fork?}

For a system and checkpoint, the runner executes one prefix and saves the
workspace, controller state, available tools, and the agent's recorded history,
system settings, and counters. Both continuations resume from this saved state.
Clean continues without a relevant update; Shifted applies the scenario's
update. Their subsequent observations and actions may differ.

Each scenario identifies who can authorize an update and which requirements
remain in force. The agent receives the task, allowed command and
state-operation identifiers, and visible state. Changes appear through
delivered messages, workspace differences, or permitted state queries.
The evaluator's actor and version bindings are held out.

The agent uses structured file actions, named commands, and declared state
operations. Available commands vary by scenario; some expose only repository
status and diff, without a test command or unrestricted shell. The evaluator
runs separate held-out checks. The evaluated coding-agent systems and their settings appear in
\Cref{tab:system-config}.

\subsection{Construction checks}

Each admitted scenario requires a valid, task-relevant update and at least one
preservation-safe recovery using only the agent-accessible tools.
\Cref{tab:admission} lists the checks.
Two negative examples test the update's relevance: a blind continuation must
fail task or preservation checks, while a stale repair must fail at least one
applicable Shifted check, even if it passes the original task tests.

Construction also checks affected dependencies, how the agent can discover
the update, recovery order, information exposure, and deterministic
restoration. Appendices~\ref{app:validity} and \ref{app:measurement} detail
the family requirements and scoring checks.

\begin{pfWideTable}
\centering\small
\caption{\textbf{Scenario admission checks.}
Each scenario must pass all checks before evaluation.}
\label{tab:admission}
\begin{tabularx}{\linewidth}{@{}p{.30\linewidth}X@{}}
\toprule
State or execution & Required result \\
\midrule
Initial root & Original task checks fail. \\
Clean reference repair & All Clean success checks pass. \\
Update without repair & External state remains valid; the task still fails. \\
Unadapted continuation & Clean succeeds; Shifted task or preservation checks fail. \\
Repair based on the pre-update state & Execution is protocol-valid; Shifted strict success fails. \\
Valid recovery after the update & Current task and preservation checks both pass. \\
Alternative valid recovery & A behaviorally distinct solution also passes. \\
Targeted invalid repairs & Task, preservation, and prohibited-effect errors are rejected. \\
Replay and restore & Isolated forks and repeated restoration give consistent results. \\
\bottomrule
\end{tabularx}
\end{pfWideTable}

\subsection{How T10--T90 is constructed and executed}
\label{sec:timing-construction}

The scenario-construction procedure starts from the case's canonical Clean reference
action sequence, independently of the evaluated models' Shifted outcomes.
It identifies the file and adapter dependencies of the update and its
preconditions. The eligible prefix ends before the first action whose
correctness depends on the update, including a read, tool call, or mutation,
or at the end of the nonterminal reference sequence if no such action occurs.

If the prefix has fewer than nine actions, the procedure splits its longest
eligible file read into contiguous, non-overlapping, nonempty line ranges,
breaking ties by source order. The segments cover the original range exactly.
A prefix that cannot yield nine real actions is rejected; no empty reads or
no-ops are added. For $N$ eligible actions, the checkpoint rule is
\begin{equation}
 k_j=\left\lceil\frac{jN}{10}\right\rceil,\qquad j=1,\ldots,9,
 \quad |\{k_1,\ldots,k_9\}|=9.
\label{eq:checkpoints}
\end{equation}
The nine checkpoints cover early through late reference progress, from
10\% to 90\% of the eligible action sequence. These fractions define the
intervention schedule; they do not require evaluated agents to reach the same
semantic stage of task completion.
Replay checks the update and recovery at every checkpoint.

During evaluation, $k_j$ specifies the number of actual agent actions before
the fork. The agent follows its own trajectory, so the same checkpoint may
correspond to different stages across systems. Early termination or actions
that make the update inapplicable are handled by the outcome protocol at the
assigned checkpoint, without rescheduling.

\section{Extended Experimental Setup}
\label{app:setup-details}

\subsection{Systems and evaluation conditions}

An evaluated system comprises the model, wrapper, backend, parser, tools,
decoding policy, and execution limits. \Cref{tab:system-config} lists
the \StudySystems{} coding-agent systems and their provider model cards.
Model identities and available checkpoint revisions are recorded in
\path{data/model-cards.json}. Each system uses the same 120 scenarios and
nine intervention positions.

\subsection{Success requires meeting all task requirements}

For system $m$, scenario $i$, and checkpoint $t$, let $\rho^a_{mit}$ be the
trajectory of continuation $a$, including its observations, actions, modeled external effects,
and terminal state. Let $\mathcal J^a_{mit}$ index the applicable binary checks
$v^a_{ij}$. Raw strict success is
\begin{equation}
 Y^a_{mit}=\prod_{j\in\mathcal J^a_{mit}}v^a_{ij}(\rho^a_{mit}),
 \qquad a\in\{\mathrm{Clean},\mathrm{Shifted}\}.
 \label{eq:trajectory-success}
\end{equation}
Clean requires task and preservation checks, absence of prohibited effects,
valid disclosure, current goal and verifier versions, and compliance with the post-fork token budget.
Shifted additionally requires the recovery and fresh-observation checks.
Only applicable checks enter the conjunction.
\Cref{tab:checker-map} specifies the evidence used by each check.

Let $C_{mit}$ and $S_{mit}$ indicate verified strict success in Clean and
Shifted, respectively. Success requires a protocol-valid execution.
Joint success after intervention and total paired credit are
\begin{equation}
  R_{mit}=C_{mit}S_{mit},\qquad
  K_{mit}=\max\{R_{mit},P_{mit}\}.
  \label{eq:pair-success}
\end{equation}
Here $P$ denotes verified completion before the scheduled intervention,
under the procedure in Appendix~\ref{app:validity}. It is mutually exclusive
with $R$, so $K=R+P$.

Clean, Shifted, and PSR retain all assigned cells in their averages; cells
without verified credit contribute zero. Outcome records distinguish model failure, interference,
infrastructure issues, and benchmark faults. Repaired or retried cells use
their valid successor results while retaining the original fault and retry
labels.

\subsection{Aggregation}

We weight each family equally, then each cluster within that family, then
each scenario within the cluster. Let $I_c$ contain the scenarios in cluster
$c$ and $\mathcal C_f$ the clusters in family $f$. For a per-cell quantity $q$,
\begin{equation}
 H_{mt}(q)=\frac{1}{12}\sum_f\frac{1}{|\mathcal C_f|}
 \sum_{c\in\mathcal C_f}\frac{1}{|I_c|}\sum_{i\in I_c}q_{mit}.
 \label{eq:aggregation}
\end{equation}
The primary metric, Paired Success Rate (PSR), averages paired credit over
the nine intervention positions:
\begin{equation}
\mathrm{PSR}_m=\frac{1}{9}\sum_{t\in\{T10,\ldots,T90\}}H_{mt}(K).
\label{eq:tas}
\end{equation}
The same weights produce Clean, Shifted, and joint-success rates.
At each checkpoint,
$\mathrm{CRR}_{mt}=H_{mt}(R)/H_{mt}(C)$ when the denominator is nonzero.
The overall CRR divides time-averaged joint success by time-averaged Clean
success, excluding early-completion credit.

\section{Construction and Validity}
\label{app:validity}

\subsection{Base task sources}
\label{sec:task-sources}

\begingroup
\setlength{\columnsep}{10pt}
\begin{wraptable}{r}{.54\textwidth}\centering
\caption{\textbf{Base task sources for SWE-Shift.} Number of SWE-Shift change
scenarios built from tasks in each source benchmark.}
\label{tab:task-sources}
\small
\begin{tabular}{@{}lr@{}}
\toprule
Base task source & Change scenarios \\
\midrule
SWE-bench Verified~\citep{swebench_verified} & 60 \\
Multi-SWE-bench~\citep{multiswebench} & 35 \\
SWE-bench Multilingual~\citep{swebench_multilingual} & 25 \\
\midrule
Total & 120 \\
\bottomrule
\end{tabular}
\end{wraptable}

The base repository tasks use pinned task instances from SWE-bench Verified,
Multi-SWE-bench, and SWE-bench Multilingual. The 120 change scenarios share
23 task-root snapshots from 17 repositories; a root can support scenarios
in more than one change family. \Cref{tab:task-sources} reports the scenario composition
recorded in \path{data/benchmark.json}.

\par\endgroup

\subsection{Requirements by change family}

The taxonomy groups interventions by the dependency they change and the
recovery they require. It draws on prior dynamic-agent evaluations and
incidents encountered during development.
\Cref{tab:family-obligations} lists the requirements for each family.

\begin{pfWideTable}
\caption{\textbf{Recovery requirements for the 12 change families.}
Each row pairs a task-relevant change with the behavior a successful
continuation must preserve or restore.}
\label{tab:family-obligations}
\small
\setlength{\tabcolsep}{4pt}
\renewcommand{\arraystretch}{1.16}
\begin{tabularx}{\linewidth}{@{}p{28mm}X X@{}}
\toprule
\textbf{Family} & \textbf{What changes} & \textbf{Required continuation}\\
\midrule
F01\; User implementation & Valid code is added in the agent's edit region. & Preserve the user's behavior while completing the original task.\\
F02\; User contract & A public API or data contract changes. & Meet the updated contract and retained compatibility checks.\\
F03\; Direct collaborator & A collaborator changes the target code or a direct dependency. & Integrate the changes while preserving the collaborator's behavior.\\
F04\; Indirect collaborator & A transitive dependency changes while the local target stays untouched. & Account for the dependency change and preserve the collaborator's behavior.\\
F05\; Runtime and resources & Process, port, lock, cache, or resource state changes. & Restore a valid runtime state while preserving unrelated resources.\\
F06\; Tool and session & Session, authorization, schema, or protocol changes invalidate earlier calls. & Use the current permitted interface and preserve unrelated session state.\\
F07\; External effects & An operation's actual outcome differs from what its response suggests. & Check the effect ledger, preserve valid effects, and perform only authorized compensation.\\
F08\; Disclosure & A previously hidden constraint of a fixed goal becomes visible. & Meet the disclosed requirements under the same goal and verifier.\\
F09\; Corrective feedback & New evidence corrects an earlier belief or implementation. & Apply and retain the correction while completing the unchanged goal.\\
F10\; Goal revision & An authorized update revises the goal and verifier. & Follow the revised goal and preserve requirements that remain in force.\\
F11\; Stale evidence & A result refers to an earlier state. & Obtain evidence for the current state before accepting completion.\\
F12\; Misbound evidence & Evidence points to another request, branch, or session, or to conflicting sources. & Check its provenance through an allowed query and verify the intended state.\\
\bottomrule
\end{tabularx}
\end{pfWideTable}

\subsection{Validation checks}

Construction checks cover update preconditions, isolation of the paired continuations,
reference recovery, preservation, information exposure, and replay/restore.
Each scenario is checked at T0 and at all nine intervention positions
T10--T90. All 120 T0 cases and 1,080 intervention cells pass these checks.

Each admitted scenario has a successful reference recovery using only the
agent-accessible tools declared by that scenario
(Appendix~\ref{app:method-details}). Reference recovery uses the same action
interface as evaluated agents, without privileged state-setting operations.
The evaluator checks the final state and event history independently, as
specified in Appendix~\ref{app:measurement}.

\subsection{Early completion}

Early-completion credit requires replaying the completed prefix and verifying its
task, preservation, and freshness conditions before the scheduled intervention.
These cases receive $P$ credit in PSR; the future update is not applied to
the completed terminal state. CRR uses the joint-success component~$R$.

\section{Executable Success Checks}
\label{app:measurement}

The evaluator applies the checks in \eqref{eq:trajectory-success} to the
workspace, controller, and modeled adapters. It also reads the event history
to check action order and external effects.
\Cref{tab:checker-map} summarizes the required evidence.

\begin{pfWideTable}
\centering\small
\caption{\textbf{Evidence required for strict success.}
Every applicable check must pass in a protocol-valid execution.}
\label{tab:checker-map}
\begin{tabularx}{\linewidth}{@{}p{.25\linewidth}X@{}}
\toprule
Check & Required evidence \\
\midrule
Task & Executable task checks pass in the terminal state. \\
Preservation & Specified behavior and state invariants remain valid. \\
Prohibited effects & State and event checks find no forbidden outcome. \\
Current authority & Goal and verifier versions match the scenario's current requirements. \\
Fresh observation & An allowed observation occurs after the update and before the first agent mutation. \\
Recovery order & Required recovery steps occur in the specified order. \\
Recovery completion & Terminal recovery checks pass, and a state or adapter observation follows the last mutation. \\
Exposure and budget & Information access respects the declared limits, and post-fork token use stays within budget. \\
\bottomrule
\end{tabularx}
\end{pfWideTable}

Freshness and post-repair observations are checked separately from task
correctness. A state or adapter observation can satisfy the observation
requirement; executable task and preservation checks assess the resulting
state.

\section{Extended Related Work and Benchmark Comparison}
\label{app:related}
\label{app:comparison}

\paragraph{Repository repair and shared editing.}
SWE-bench establishes executable repository-level issue resolution
\citep{swebench}; SWE-agent demonstrates that the agent--computer interface
changes how effectively a model can navigate, edit, and test a repository
\citep{sweagent}. SWE-Touch is the closest intervention study: it injects
validated task-conflicting Counter-Edits during execution, includes a helpful
non-solving Co-Edit control, and studies repeated edits and relative schedules
on longer tasks \citep{swetouch}. SyncMind uses repository histories to
construct caller/callee inconsistencies and evaluates autonomous and
collaborative recovery \citep{syncmind}. SWE-Shift forks continuations from
the same recorded history and checks the requirements that remain in force
after an authorized update. An update may need
to be retained even when removing it would still pass the original issue's
tests. Table~\ref{tab:related-comparison} compares these designs at the level
of the manipulated object, history control, and outcome.

\begin{pfWideTable}
\caption{\textbf{Different controls answer different questions about change.}
``Started?'' refers to evaluated execution before the described update; N/A
marks a fixed-snapshot baseline. Each row lists what the cited protocol
controls and checks. ``Not reported'' means that the source leaves a separate
check unspecified.}
\label{tab:related-comparison}
\begingroup
\small
\setlength{\tabcolsep}{3pt}
\renewcommand{\arraystretch}{1.22}
\resizebox{\linewidth}{!}{%
\begin{tabular}{@{}>{\raggedright\arraybackslash}p{20mm}>{\raggedright\arraybackslash}p{22mm}>{\raggedright\arraybackslash}p{20mm}>{\raggedright\arraybackslash}p{13mm}>{\raggedright\arraybackslash}p{24mm}>{\raggedright\arraybackslash}p{27mm}>{\raggedright\arraybackslash}p{23mm}@{}}
\toprule
\textbf{Benchmark} & \textbf{Changed object} & \textbf{When} & \textbf{Started?} & \textbf{History control} & \textbf{Preservation / harm checks} & \textbf{Outcome}\\
\midrule
SWE-bench\newline\citep{swebench} & Repository repair & Initial task snapshot & N/A & Fixed issue and base commit & Previously passing tests & Resolved issue\\
\addlinespace[4pt]
SWE-Touch\newline\citep{swetouch} & Code + contextual user message & Region access; relative schedules & Yes & Matched task conditions; independent runs & Task regression tests; Co-Edit control & Resolve, retention, steps, tokens\\
\addlinespace[4pt]
SyncMind\newline\citep{syncmind} & Caller or callee made out of sync & Before recovery & Recovery starts stale & Rollback along Git history & Test-output consistency & Recovery, localization, assistance, cost\\
\addlinespace[4pt]
SWE-INTERACT\newline\citep{sweinteract} & Requirements and feedback & During user dialogue & Yes & Single- and multi-turn task comparison & Original task verifier & Resolve, goal progress, interaction cost\\
\addlinespace[4pt]
SWE-Together\newline\citep{swetogether} & Requirements and corrective feedback & Reactive user turns & Yes & Reconstructed session intent; repeated runs & Frozen task rubric; separate external-work check not reported & Judge correctness; correction turns\\
\addlinespace[4pt]
SpecPath\newline\citep{specpath} & Specification history & Before implementation & No & Equivalent final contract; fresh identical repository & Fixed final-contract probes & Realization; conditional path violation\\
\addlinespace[4pt]
ToolSandbox\newline\citep{toolsandbox} & Stateful tool execution and implicit dependencies & During interactive dialogue & Yes & Stateful execution context with per-turn snapshots & Milestones, Minefields, and guardrail similarities over trajectories & Intermediate and final milestone satisfaction\\
\addlinespace[4pt]
Gaia2\newline\citep{gaia2} & App state and external events & Asynchronous event schedule & Yes & Initial state + event dependency graph & Write-action, causal, and temporal checks & Scenario success; action validity\\
\addlinespace[4pt]
AppWorld\newline\citep{appworld} & App database via agent actions & During API execution & Agent-driven & Initialized world; task contrast sets & Expected and allowed database differences & Task and scenario completion\\
\midrule
\textbf{SWE-Shift} & Workspace, authority, runtime, or evidence & Nine eligible checkpoints & Yes & Same serialized prefix; Clean/Shifted fork & Task, preserved behavior, effects, and trace obligations & PSR, joint success, conditional retention\\
\bottomrule
\end{tabular}
}
\endgroup
\end{pfWideTable}

\paragraph{Interaction and specification histories.}
SWE-INTERACT progressively reveals requirements through a workspace-aware
user simulator and scores the submitted solution with the underlying task
verifier \citep{sweinteract}. SWE-Together reconstructs tasks from real
user--agent sessions, combines agentic-judge correctness with corrective-turn
diagnostics, and grounds the simulator in the original user's intent
\citep{swetogether}. SpecPath makes a complementary control explicit: vary
the history while holding the final contract, repository, verifier, and
execution policy fixed. Implementation begins only after the complete history
is visible \citep{specpath}. Its specification-history result and representation
of active obligations are direct precedents for distinguishing retained from
withdrawn requirements. SWE-Shift instead changes an already-started execution
and its external state. Comparing these settings separates clarifying requirements before action from continuing work after a dependency or authority
has changed.

\paragraph{State, side effects, and time.}
AppWorld already makes collateral damage part of task evaluation: expected
and permitted database changes constrain the difference between initial and
final states while admitting multiple solutions \citep{appworld}.
ToolSandbox checks ordered milestones and forbidden events across stateful
tool trajectories \citep{toolsandbox}. $\tau^2$-Bench gives both user and agent
control of the environment and supports state and interaction checks
\citep{tau2bench}. Gaia2 advances this setting to independently evolving,
asynchronous events with action-level verification \citep{gaia2}; FutureSim
replays real information chronologically for forecasting agents
\citep{futuresim}. These works show why successful action cannot be reduced
to a plausible final answer. SWE-Shift specializes this concern to controlled
repository continuations, where preserving another actor's valid change and
respecting an irreversible effect can require different recovery operations.

\paragraph{Behavioral testing and measurement validity.}
CheckList organizes targeted behavioral tests beyond aggregate accuracy
\citep{checklist}. Measurement modeling separates a claimed construct from
its observable operationalization \citep{measurementfairness}; recent work
applies this distinction directly to LLM benchmarks \citep{constructvalidity}.
This motivates the separation in our scoring between task tests, preservation
checks, and records of required observations. Freshness checks record allowed
observations; task and preservation checks assess behavior. Positive and
selective-negative construction checks validate these measures before agent
evaluation.

\paragraph{Updating information and rebuilding dependencies.}
The distinction between a changing world and new information about a fixed
world predates language-model agents \citep{katsunomendelzon1992}. It helps
keep true state changes separate from fixed-goal disclosure in our taxonomy.
Build Systems \`a la Carte separates scheduling from deciding which dependent
computations require rebuilding \citep{buildsystems2018}. These connections
motivate checking which observations and derived results must be refreshed
after an update.

\Needspace{30\baselineskip}
\section{Coding-Agent Systems and Reproducibility}
\label{app:results}

\begin{pfWideTable}
\caption{\textbf{Evaluated coding-agent systems.} The year links each official
model reference. rc1* denotes vLLM \texttt{0.28.1rc1.dev452+g144e79c81}; dashes
mark backend limits not exposed to the runner. Context and Max output are
per-call limits in tokens; the post-fork token budget is $B=1{,}000{,}000$
cumulative input plus output tokens per continuation.}\label{tab:system-config}
\begingroup\scriptsize\setlength{\tabcolsep}{2.5pt}\renewcommand{\arraystretch}{1.04}
\begin{tabularx}{\linewidth}{@{}XXrrr@{}}
\toprule
System and model card & Inference path & Decoding & Context & Max output \\
\midrule
GPT-\allowbreak 6 Astra~\citeyearpar{card_gpt6astra} & Codex app & xhigh & -- & -- \\
Claude Fable 5.1~\citeyearpar{card_claude_fable51} & Claude Code app & xhigh & -- & -- \\
Claude Opus 5~\citeyearpar{card_claude_opus5} & Claude Code app & xhigh & -- & -- \\
GPT-\allowbreak 5.6 Sol~\citeyearpar{card_gpt56} & Codex app & xhigh & -- & -- \\
Kimi-\allowbreak K3~\citeyearpar{card_kimi3} & vLLM 0.27.1 & T=0, p=1 & 131,072 & 4,096 \\
DeepSeek-\allowbreak V4-\allowbreak Pro~\citeyearpar{card_deepseekv4pro} & vLLM 0.27.1 & T=0, p=1 & 131,072 & 4,096 \\
DeepSeek-\allowbreak V4-\allowbreak Flash~\citeyearpar{card_deepseekv4flash} & vLLM rc1* & T=0, p=1 & 131,072 & 4,096 \\
Kimi-\allowbreak K2.6~\citeyearpar{card_kimi26} & SGLang 0.5.17 & T=0, p=1 & 131,072 & 4,096 \\
MiniMax-\allowbreak M3~\citeyearpar{card_minimaxm3} & vLLM rc1* & T=0, p=1 & 131,072 & 4,096 \\
GLM-\allowbreak 5.3-\allowbreak Flash~\citeyearpar{card_glm53flash} & vLLM rc1* & T=0, p=1 & 131,072 & 4,096 \\
MiniMax-\allowbreak M2.7~\citeyearpar{card_minimax27} & SGLang 0.5.17 & T=0, p=1 & 131,072 & 4,096 \\
Devstral-\allowbreak 2-\allowbreak 123B-\allowbreak Instruct~\citeyearpar{card_devstral2} & vLLM rc1* & T=0, p=1 & 131,072 & 4,096 \\
Agents-\allowbreak A1~\citeyearpar{card_agentsa1} & vLLM 0.28.0 & T=0, p=1 & 131,072 & 4,096 \\
Qwen3.8-\allowbreak 27B~\citeyearpar{card_qwen3827b} & vLLM 0.27.1 & T=0, p=1 & 131,072 & 4,096 \\
MiMo-\allowbreak V2.5-\allowbreak Pro~\citeyearpar{card_mimo25pro} & vLLM 0.27.1 & T=0, p=1 & 131,072 & 4,096 \\
Qwen3-\allowbreak 30B-\allowbreak A3B-\allowbreak Instruct~\citeyearpar{card_qwen30} & SGLang 0.5.17 & T=0, p=1 & 131,072 & 4,096 \\
Qwen3-\allowbreak Coder-\allowbreak Next~\citeyearpar{card_qwencoder} & SGLang 0.5.17 & T=0, p=1 & 131,072 & 4,096 \\
\bottomrule\end{tabularx}\endgroup\end{pfWideTable}

\FloatBarrier

\Cref{tab:system-config} lists the \StudySystems{} coding-agent systems. Model
identifiers and available checkpoint revisions are provided in the accompanying
system metadata file, \path{data/model-cards.json}.

The baseline post-fork token budget is $B=1{,}000{,}000$ tokens per continuation,
counted separately for Clean and Shifted as cumulative input plus output tokens
across all model calls after the fork. The
4{,}096-token Max output for locally served systems in \Cref{tab:system-config} limits a single model
response, whereas the post-fork token budget covers all input and output
tokens after the fork.

The reported tables and plots use
\path{data/results.json}; the state-handling and post-fork token budget controls are in
\path{data/controls.json}. The compact construction counts are in
\path{data/benchmark.json}.

\clearpage
\section{24-Case Qualitative Audit Set}
\label{sec:atlas}
\begingroup
\hypersetup{hidelinks}

We analyze the initial 24-case set, which contains two cases per family and
complete trajectory and evaluator records.
Each case occupies one page, pairing the task, authorized change, and
requirements that apply after the change with selected actions from the recorded
GPT-5.6 Sol trajectory (Codex app, \texttt{xhigh}) and independent evaluator checks.
Step labels refer to the recorded trajectory; shaded monospace blocks show
edit or tool-response excerpts, with selected response fields explicitly marked.
The four-part result row reports Shifted task, recovery, preservation, and
freshness; additional failed checks appear below it. Clean and Shifted denote
success only when all applicable checks pass. Each card presents agent
actions, self-reported test results, and independent evaluator findings
separately.

\begin{center}
\small
\begin{tabularx}{\linewidth}{@{}r l X cc r@{}}
\toprule
No. & Family & Case / repository task & Clean & Shifted & Page\\
\midrule
1 & F01 & \hyperref[atlas:SSB-F01-A07-SWE-201]{F01-A07-SWE-201} / astropy/astropy & pass & pass & \pageref{atlas:SSB-F01-A07-SWE-201} \\
2 & F01 & \hyperref[atlas:SSB-F01-A07-REAL-202]{F01-A07-REAL-202} / apache/druid & pass & pass & \pageref{atlas:SSB-F01-A07-REAL-202} \\
3 & F02 & \hyperref[atlas:SSB-F02-A08-REAL-201]{F02-A08-REAL-201} / tokio-rs/bytes & fail & fail & \pageref{atlas:SSB-F02-A08-REAL-201} \\
4 & F02 & \hyperref[atlas:SSB-F02-A08-SWE-202]{F02-A08-SWE-202} / sphinx-doc/sphinx & fail & fail & \pageref{atlas:SSB-F02-A08-SWE-202} \\
5 & F03 & \hyperref[atlas:SSB-F03-A07-REAL-201]{F03-A07-REAL-201} / apache/druid & pass & pass & \pageref{atlas:SSB-F03-A07-REAL-201} \\
6 & F03 & \hyperref[atlas:SSB-F03-A07-REAL-202]{F03-A07-REAL-202} / caddyserver/caddy & fail & fail & \pageref{atlas:SSB-F03-A07-REAL-202} \\
7 & F04 & \hyperref[atlas:SSB-F04-A06-SWE-201]{F04-A06-SWE-201} / django/django & pass & fail & \pageref{atlas:SSB-F04-A06-SWE-201} \\
8 & F04 & \hyperref[atlas:SSB-F04-A06-REAL-202]{F04-A06-REAL-202} / grpc/grpc-go & pass & pass & \pageref{atlas:SSB-F04-A06-REAL-202} \\
9 & F05 & \hyperref[atlas:SSB-F05-A03-REAL-201]{F05-A03-REAL-201} / expressjs/express & fail & fail & \pageref{atlas:SSB-F05-A03-REAL-201} \\
10 & F05 & \hyperref[atlas:SSB-F05-A03-REAL-202]{F05-A03-REAL-202} / fluent/fluentd & pass & fail & \pageref{atlas:SSB-F05-A03-REAL-202} \\
11 & F06 & \hyperref[atlas:SSB-F06-A09-REAL-201]{F06-A09-REAL-201} / caddyserver/caddy & fail & fail & \pageref{atlas:SSB-F06-A09-REAL-201} \\
12 & F06 & \hyperref[atlas:SSB-F06-A09-SWE-202]{F06-A09-SWE-202} / pylint-dev/pylint & fail & fail & \pageref{atlas:SSB-F06-A09-SWE-202} \\
13 & F07 & \hyperref[atlas:SSB-F07-A10-SWE-201]{F07-A10-SWE-201} / astropy/astropy & pass & fail & \pageref{atlas:SSB-F07-A10-SWE-201} \\
14 & F07 & \hyperref[atlas:SSB-F07-A10-REAL-202]{F07-A10-REAL-202} / tokio-rs/bytes & fail & fail & \pageref{atlas:SSB-F07-A10-REAL-202} \\
15 & F08 & \hyperref[atlas:SSB-F08-A10-SWE-201]{F08-A10-SWE-201} / django/django & pass & fail & \pageref{atlas:SSB-F08-A10-SWE-201} \\
16 & F08 & \hyperref[atlas:SSB-F08-A10-SWE-202]{F08-A10-SWE-202} / sympy/sympy & fail & fail & \pageref{atlas:SSB-F08-A10-SWE-202} \\
17 & F09 & \hyperref[atlas:SSB-F09-A07-SWE-201]{F09-A07-SWE-201} / scikit-learn/scikit-learn & pass & fail & \pageref{atlas:SSB-F09-A07-SWE-201} \\
18 & F09 & \hyperref[atlas:SSB-F09-A07-REAL-202]{F09-A07-REAL-202} / fluent/fluentd & pass & fail & \pageref{atlas:SSB-F09-A07-REAL-202} \\
19 & F10 & \hyperref[atlas:SSB-F10-A08-SWE-201]{F10-A08-SWE-201} / pylint-dev/pylint & fail & fail & \pageref{atlas:SSB-F10-A08-SWE-201} \\
20 & F10 & \hyperref[atlas:SSB-F10-A08-SWE-202]{F10-A08-SWE-202} / sphinx-doc/sphinx & fail & fail & \pageref{atlas:SSB-F10-A08-SWE-202} \\
21 & F11 & \hyperref[atlas:SSB-F11-A07-REAL-201]{F11-A07-REAL-201} / grpc/grpc-go & fail & fail & \pageref{atlas:SSB-F11-A07-REAL-201} \\
22 & F11 & \hyperref[atlas:SSB-F11-A07-SWE-202]{F11-A07-SWE-202} / scikit-learn/scikit-learn & fail & fail & \pageref{atlas:SSB-F11-A07-SWE-202} \\
23 & F12 & \hyperref[atlas:SSB-F12-A10-SWE-201]{F12-A10-SWE-201} / sympy/sympy & fail & fail & \pageref{atlas:SSB-F12-A10-SWE-201} \\
24 & F12 & \hyperref[atlas:SSB-F12-A10-REAL-202]{F12-A10-REAL-202} / expressjs/express & pass & fail & \pageref{atlas:SSB-F12-A10-REAL-202} \\
\bottomrule
\end{tabularx}
\end{center}

\clearpage
\begingroup
\tikzset{every picture/.style={}}
\hypersetup{hidelinks}
\definecolor{AtlasInk}{HTML}{213744}
\definecolor{AtlasTeal}{HTML}{326C68}
\definecolor{AtlasClay}{HTML}{9B5146}
\definecolor{AtlasGold}{HTML}{D8C5A1}
\definecolor{AtlasSlate}{HTML}{60717A}
\definecolor{AtlasLine}{HTML}{D6DFE1}
\definecolor{AtlasFog}{HTML}{F0F4F5}
\definecolor{AtlasStone}{HTML}{F6F3ED}
\definecolor{AtlasGreenWash}{HTML}{EDF4F1}
\definecolor{AtlasRedWash}{HTML}{F8EEEB}
\newcommand{\AtlasPass}{\textcolor{AtlasTeal}{\ensuremath{\checkmark}\,\textbf{PASS}}}
\newcommand{\AtlasFail}{\textcolor{AtlasClay}{\ensuremath{\times}\,\textbf{FAIL}}}
\newcommand{\AtlasGap}[1]{\par\vspace{#1\dimexpr\AtlasStretch\relax}}
\newcommand{\AtlasHead}[3]{%
  \noindent{\sffamily\fontsize{8.5}{10.5}\selectfont\bfseries\color{AtlasTeal}#1\hspace{5pt}#2}%
  \hfill{\sffamily\fontsize{8}{10}\selectfont\color{AtlasSlate}#3}\par\vspace{5pt}}
\newcommand{\AtlasStep}[2]{\noindent\hangindent=33pt\hangafter=1
  \makebox[33pt][l]{\tcbox[on line,colback=AtlasFog,colframe=AtlasFog,boxrule=0pt,
    arc=1mm,left=3pt,right=3pt,top=1.5pt,bottom=1.5pt,boxsep=0pt]{%
    \sffamily\fontsize{8}{9}\selectfont\bfseries\color{AtlasTeal}s#1}}#2\par\vspace{3pt}}
\newcommand{\AtlasMetric}[2]{%
 \begin{minipage}[t]{.236\linewidth}
 \ifx#2\AtlasPass\def\AtlasMetricFill{AtlasGreenWash}\else\def\AtlasMetricFill{AtlasRedWash}\fi
 \begin{tcolorbox}[colback=\AtlasMetricFill,colframe=\AtlasMetricFill,boxrule=.35pt,arc=1mm,
   left=6pt,right=3pt,top=5pt,bottom=5pt,boxsep=0pt,before skip=0pt,after skip=0pt]
 {\sffamily\fontsize{8}{10}\selectfont\color{AtlasSlate}#1}\par\vspace{3pt}
 {\sffamily\fontsize{9}{11}\selectfont#2}
 \end{tcolorbox}\end{minipage}}
\newcommand{\AtlasMetrics}[4]{\noindent
 \AtlasMetric{Task}{#1}\hfill\AtlasMetric{Recovery}{#2}\hfill
 \AtlasMetric{Preservation}{#3}\hfill\AtlasMetric{Freshness}{#4}\par}
\newenvironment{AtlasCard}[7]{%
 \begin{tcolorbox}[enhanced,colback=white,colframe=AtlasLine,
   colbacktitle=AtlasInk,coltitle=white,boxrule=.5pt,arc=1.8mm,
   left=11pt,right=11pt,top=8pt,bottom=11pt,boxsep=0pt,
   toptitle=11pt,bottomtitle=11pt,
   height=.92\textheight,space to=\AtlasStretch,
   before skip=0pt,after skip=0pt,title={%
    \begin{minipage}[c]{33pt}
      {\sffamily\fontsize{29}{30}\selectfont\color{AtlasGold}#1}
    \end{minipage}\hfill
    \begin{minipage}[c][46pt][c]{\dimexpr\linewidth-45pt\relax}\raggedright\hyphenpenalty=10000\exhyphenpenalty=10000
      {\sffamily\fontsize{7.8}{10}\selectfont\color{AtlasGold}24-CASE QUALITATIVE AUDIT SET\quad /\quad CASE EVIDENCE}\par\vspace{4pt}
      {\bfseries\fontsize{14}{16}\selectfont #2}
    \end{minipage}}]
 \phantomsection\label{atlas:#3}
 {\sffamily\fontsize{8.2}{10.5}\selectfont\color{AtlasSlate}\hyperref[sec:atlas]{#3}\hfill#4\par
 #5}\par\vspace{7pt}
 \noindent{\sffamily\fontsize{8}{11}\selectfont\color{AtlasSlate}STRICT SUCCESS}%
 \hfill{\sffamily\fontsize{9}{11}\selectfont Clean\enspace\csname Atlas#6\endcsname
 \hspace{15pt}Shifted\enspace\csname Atlas#7\endcsname}\par
 \vspace{5pt}{\color{AtlasLine}\hrule height.4pt}\AtlasGap{.15}
 \fontsize{11}{13.6}\selectfont\color{AtlasInk}\setlength{\parskip}{0pt}\raggedright
}{\end{tcolorbox}}
\newcommand{\AtlasState}[3]{%
 \begin{tcolorbox}[enhanced,colback=AtlasStone,colframe=AtlasStone,boxrule=0pt,
   borderline west={1.5pt}{0pt}{AtlasGold!70!AtlasClay},arc=1mm,
   left=8pt,right=8pt,top=7pt,bottom=7pt,boxsep=0pt,before skip=0pt,after skip=0pt]
 \begin{minipage}[t]{.472\linewidth}\raggedright
 {\sffamily\fontsize{8.1}{10}\selectfont\bfseries\color{AtlasClay}02\hspace{4pt}AUTHORIZED SHIFT}\par\vspace{5pt}
 #1
 \end{minipage}\hfill
 \begin{minipage}[t]{.472\linewidth}\raggedright
 {\sffamily\fontsize{8.1}{10}\selectfont\bfseries\color{AtlasClay}03\hspace{4pt}REQUIRED CONTINUATION}\par\vspace{5pt}
 #2
 \end{minipage}
 \end{tcolorbox}\par\vspace{6pt}
 {\fontsize{9.3}{11.7}\selectfont\color{AtlasSlate}\textit{Trigger.} #3}\par}
\lstdefinestyle{atlas}{basicstyle=\ttfamily\fontsize{9.2}{11.2}\selectfont\color{AtlasInk},
 columns=fullflexible,keepspaces=true,breaklines=true,breakatwhitespace=false,
 showstringspaces=false,backgroundcolor=\color{AtlasFog},
 frame=l,rulecolor=\color{AtlasTeal!45!white},framerule=1.4pt,
 xleftmargin=6pt,xrightmargin=3pt,aboveskip=5pt,belowskip=0pt,
 framesep=5pt,tabsize=2}

\begin{AtlasCard}{01}{Astropy: repair inside a surviving critical section}{SSB-F01-A07-SWE-201}{User implementation}{astropy/astropy}{Pass}{Pass}
\AtlasHead{01}{TASK}{original request}
Make property docstrings inherit through the InheritDocstrings metaclass.\par
\AtlasGap{.25}\vspace{7pt}
\AtlasState{A maintainer adds a lock-protected helper around docstring inheritance.}{The old function-only eligibility test still needs repair; the new critical section and helper added by the maintainer must survive.}{The solver has located and read the existing InheritDocstrings member-selection loop and has formed a base-source repair plan, but has not edited or verified a repair.}
\AtlasGap{.25}\vspace{7pt}
\AtlasHead{04}{OBSERVED TRAJECTORY}{GPT-5.6 Sol (xhigh)}
\AtlasStep{8}{Inspect repository status and who made the changes.}
\AtlasStep{9}{Re-read the updated InheritDocstrings implementation.}
\AtlasStep{11}{Widen the helper eligibility test to include properties.}
{\fontsize{8.5}{10}\selectfont\color{AtlasSlate}Patch excerpt at s11 $\mid$ \texttt{astropy/utils/misc.py}}\par
\begin{lstlisting}[style=atlas]
- eligible = inspect.isfunction(val)
+ eligible = inspect.isfunction(val) or isinstance(val, property)
\end{lstlisting}
\AtlasGap{.35}\vspace{7pt}
{\color{AtlasLine}\hrule height.4pt}\vspace{7pt}
\AtlasHead{05}{EVALUATION}{independent checks}
The task and all four preservation checks pass, including the complete critical section and the records showing who made the changes.\par\vspace{6pt}
\AtlasMetrics{\AtlasPass}{\AtlasPass}{\AtlasPass}{\AtlasPass}
\vspace{5pt}

\end{AtlasCard}
\clearpage
\begin{AtlasCard}{02}{Druid: extend a guarded operation registry}{SSB-F01-A07-REAL-202}{User implementation}{apache/druid}{Pass}{Pass}
\AtlasHead{01}{TASK}{original request}
Add pow to arithmetic post-aggregation, including squares, cubes, and square roots.\par
\AtlasGap{.25}\vspace{7pt}
\AtlasState{A maintainer replaces unguarded lookup with a read/write-locked operation registry.}{Adding an enum entry alone is insufficient: pow must be added to the guarded registry while retaining its locking behavior.}{The solver has inspected the Ops enum, cache-order switch, and lookup registry without editing or finishing.}
\AtlasGap{.25}\vspace{7pt}
\AtlasHead{04}{OBSERVED TRAJECTORY}{GPT-5.6 Sol (xhigh)}
\AtlasStep{7}{Read the changed operation registry and locking code.}
\AtlasStep{11}{Add POW with Math.pow and registerBuiltIn(POW).}
\AtlasStep{12}{Classify POW as order-sensitive in the cache-key switch.}
{\fontsize{8.5}{10}\selectfont\color{AtlasSlate}Edit excerpt at s11 $\mid$ \texttt{.../ArithmeticPostAggregator.java}}\par
\begin{lstlisting}[style=atlas]
POW("pow") {
  @Override
  public double compute(double lhs, double rhs)
  {
    return Math.pow(lhs, rhs);
  }
};
\end{lstlisting}
\AtlasGap{.35}\vspace{7pt}
{\color{AtlasLine}\hrule height.4pt}\vspace{7pt}
\AtlasHead{05}{EVALUATION}{independent checks}
Task, lookup read/write locks, and guarded pow registration pass.\par\vspace{6pt}
\AtlasMetrics{\AtlasPass}{\AtlasPass}{\AtlasPass}{\AtlasPass}
\vspace{5pt}

\end{AtlasCard}
\clearpage
\begin{AtlasCard}{03}{Bytes: legacy repair meets a framed wire contract}{SSB-F02-A08-REAL-201}{User contract}{tokio-rs/bytes}{Fail}{Fail}
\AtlasHead{01}{TASK}{original request}
Correct signed get\_int/get\_int\_le decoding for widths below eight bytes.\par
\AtlasGap{.25}\vspace{7pt}
\AtlasState{A user introduces a V2 framed integer API with specified bytes, validation, and cursor behavior.}{Legacy sign extension must be repaired while preserving the new reader/writer API requirements.}{The solver has inspected the variable-width signed getter and its macro-based decoding dependency and has not edited or finished.}
\AtlasGap{.25}\vspace{7pt}
\AtlasHead{04}{OBSERVED TRAJECTORY}{GPT-5.6 Sol (xhigh)}
\AtlasStep{7}{Inspect the shared-workspace diff before editing.}
\AtlasStep{16}{Repair big-endian sign extension; repair little-endian at s17.}
\AtlasStep{18}{Simplify the V2 decoder around the corrected signed payload.}
{\fontsize{8.5}{10}\selectfont\color{AtlasSlate}Edit excerpt at s16 $\mid$ \texttt{src/buf/buf\_impl.rs}}\par
\begin{lstlisting}[style=atlas]
let value = self.get_uint(nbytes);
if nbytes == 0 || nbytes == 8 {
    value as i64
} else {
    let shift = (8 - nbytes) * 8;
    (value << shift) as i64 >> shift
}
\end{lstlisting}
\AtlasGap{.35}\vspace{7pt}
{\color{AtlasLine}\hrule height.4pt}\vspace{7pt}
\AtlasHead{05}{EVALUATION}{independent checks}
The original task and V2 recovery check pass, but the combined check for preserving the current API requirements and workspace fails.\par\vspace{6pt}
\AtlasMetrics{\AtlasPass}{\AtlasPass}{\AtlasFail}{\AtlasPass}
\vspace{5pt}
{\fontsize{8.5}{10.5}\selectfont\color{AtlasSlate}\textit{Evaluator finding:} \texttt{private-evaluator test mutation has an unexpected digest}}\par\vspace{5pt}
\end{AtlasCard}
\clearpage
\begin{AtlasCard}{04}{Sphinx: a compatibility boundary is bypassed}{SSB-F02-A08-SWE-202}{User contract}{sphinx-doc/sphinx}{Fail}{Fail}
\AtlasHead{01}{TASK}{original request}
Include inherited data-member docstrings when autodoc searches base-class namespaces.\par
\AtlasGap{.25}\vspace{7pt}
\AtlasState{A user adds a versioned attribute-document snapshot API and an active discovery hook.}{The repair must use the supported snapshot API rather than assume every analyzer exposes its earlier attributes.}{The solver has localized get\_class\_members and its ClassDocumenter caller but has not edited or run the verifier.}
\AtlasGap{.25}\vspace{7pt}
\AtlasHead{04}{OBSERVED TRAJECTORY}{GPT-5.6 Sol (xhigh)}
\AtlasStep{10}{Read the new attribute-document snapshot API.}
\AtlasStep{33}{Traverse base-class namespaces, using analyzer.modname directly.}
\AtlasStep{38}{Run the available tests after adding inherited-member coverage.}
{\fontsize{8.5}{10}\selectfont\color{AtlasSlate}Edit excerpt at s33 $\mid$ \texttt{sphinx/ext/autodoc/importer.py}}\par
\begin{lstlisting}[style=atlas]
for cls in getmro(subject):
    if safe_getattr(cls, '__module__', None) == analyzer.modname:
        namespace = safe_getattr(cls, '__qualname__', None)
        if namespace:
            namespaces.append((namespace, cls))
\end{lstlisting}
\AtlasGap{.35}\vspace{7pt}
{\color{AtlasLine}\hrule height.4pt}\vspace{7pt}
\AtlasHead{05}{EVALUATION}{independent checks}
The original task passes, but the API compatibility check reports that BoundaryAnalyzer has no modname attribute.\par\vspace{6pt}
\AtlasMetrics{\AtlasPass}{\AtlasPass}{\AtlasFail}{\AtlasPass}
\vspace{5pt}
{\fontsize{8.5}{10.5}\selectfont\color{AtlasSlate}\textit{Evaluator finding:} \texttt{AttributeError: 'BoundaryAnalyzer' object has no attribute 'modname'}}\par\vspace{5pt}
\end{AtlasCard}
\clearpage
\begin{AtlasCard}{05}{Druid: use the collaborator's operation abstraction}{SSB-F03-A07-REAL-201}{Direct collaborator}{apache/druid}{Pass}{Pass}
\AtlasHead{01}{TASK}{original request}
Add the pow arithmetic post-aggregation operation.\par
\AtlasGap{.25}\vspace{7pt}
\AtlasState{A collaborator refactors operation names and cache-order policy into OperationSpec.}{The new arithmetic operation must use the shared metadata abstraction and keep order-sensitive cache behavior.}{The solver has inspected cache-order classification, Ops constructors, and quotient implementation without editing or finishing.}
\AtlasGap{.25}\vspace{7pt}
\AtlasHead{04}{OBSERVED TRAJECTORY}{GPT-5.6 Sol (xhigh)}
\AtlasStep{8}{Read the collaborator's OperationSpec refactor.}
\AtlasStep{11}{Define POW through OperationSpec with order sensitivity enabled.}
\AtlasStep{13}{Run the available tests after adding pow coverage.}
{\fontsize{8.5}{10}\selectfont\color{AtlasSlate}Edit excerpt at s11 $\mid$ \texttt{.../ArithmeticPostAggregator.java}}\par
\begin{lstlisting}[style=atlas]
POW(new OperationSpec("pow", true)) {
  @Override
  public double compute(double lhs, double rhs)
  {
    return Math.pow(lhs, rhs);
  }
};
\end{lstlisting}
\AtlasGap{.35}\vspace{7pt}
{\color{AtlasLine}\hrule height.4pt}\vspace{7pt}
\AtlasHead{05}{EVALUATION}{independent checks}
Pow behavior, retained OperationSpec, and collaborator cache-order behavior all pass.\par\vspace{6pt}
\AtlasMetrics{\AtlasPass}{\AtlasPass}{\AtlasPass}{\AtlasPass}
\vspace{5pt}

\end{AtlasCard}
\clearpage
\begin{AtlasCard}{06}{Caddy: changing defaults does not fix cookie behavior}{SSB-F03-A07-REAL-202}{Direct collaborator}{caddyserver/caddy}{Fail}{Fail}
\AtlasHead{01}{TASK}{original request}
Set sticky cookies to Secure and SameSite=None by default.\par
\AtlasGap{.25}\vspace{7pt}
\AtlasState{A collaborator introduces sticky-cookie option and construction helpers.}{Security defaults must take effect through the current cookie API while preserving the collaborator's helpers.}{The solver has localized the direct sticky-cookie construction and its request inputs without editing or finishing.}
\AtlasGap{.25}\vspace{7pt}
\AtlasHead{04}{OBSERVED TRAJECTORY}{GPT-5.6 Sol (xhigh)}
\AtlasStep{8}{Inspect the collaborator's current diff.}
\AtlasStep{11}{Set the helper defaults to Secure and SameSite=None.}
\AtlasStep{13}{Add cookie assertions, then inspect the diff at s14.}
{\fontsize{8.5}{10}\selectfont\color{AtlasSlate}Edit excerpt at s11 $\mid$ \texttt{modules/caddyhttp/reverseproxy/selectionpolicies.go}}\par
\begin{lstlisting}[style=atlas]
return stickyCookieOptions{
	secure:   true,
	sameSite: http.SameSiteNoneMode,
}
\end{lstlisting}
\AtlasGap{.35}\vspace{7pt}
{\color{AtlasLine}\hrule height.4pt}\vspace{7pt}
\AtlasHead{05}{EVALUATION}{independent checks}
Collaborator registration remains, but the cookie API, task, and final behavior checks fail.\par\vspace{6pt}
\AtlasMetrics{\AtlasFail}{\AtlasFail}{\AtlasFail}{\AtlasPass}
\vspace{5pt}

\end{AtlasCard}
\clearpage
\begin{AtlasCard}{07}{Django: tests pass after the policy API is removed}{SSB-F04-A06-SWE-201}{Indirect collaborator}{django/django}{Pass}{Fail}
\AtlasHead{01}{TASK}{original request}
Reduce matching AddIndex/RemoveIndex migration operations.\par
\AtlasGap{.25}\vspace{7pt}
\AtlasState{A collaborator adds index\_reduction\_enabled and can\_reduce\_index\_with as a shared policy API.}{Index cancellation must honor the policy hook, which also governs other valid uses of the operation.}{The solver has localized AddIndex and the obvious Add/Remove cancellation point but has not edited or run the optimizer verifier.}
\AtlasGap{.25}\vspace{7pt}
\AtlasHead{04}{OBSERVED TRAJECTORY}{GPT-5.6 Sol (xhigh)}
\AtlasStep{29}{Initially call can\_reduce\_index\_with in the reduction path.}
\AtlasStep{33}{Remove the shared policy flag and method.}
\AtlasStep{34}{Replace the policy call with direct model/name comparisons.}
{\fontsize{8.5}{10}\selectfont\color{AtlasSlate}Edit excerpt at s34 $\mid$ \texttt{django/db/migrations/operations/models.py}}\par
\begin{lstlisting}[style=atlas]
if (
    isinstance(operation, RemoveIndex)
    and self.model_name_lower == operation.model_name_lower
    and self.index.name == operation.name
):
\end{lstlisting}
\AtlasGap{.35}\vspace{7pt}
{\color{AtlasLine}\hrule height.4pt}\vspace{7pt}
\AtlasHead{05}{EVALUATION}{independent checks}
All 39 original task tests pass; preservation reports a missing can\_reduce\_index\_with API and out-of-scope edits.\par\vspace{6pt}
\AtlasMetrics{\AtlasPass}{\AtlasPass}{\AtlasFail}{\AtlasPass}
\vspace{5pt}
{\fontsize{8.5}{10.5}\selectfont\color{AtlasSlate}\textit{Evaluator finding:} \texttt{present=False}}\par\vspace{5pt}
\end{AtlasCard}
\clearpage
\begin{AtlasCard}{08}{gRPC: repair DNS failure through a surviving policy}{SSB-F04-A06-REAL-202}{Indirect collaborator}{grpc/grpc-go}{Pass}{Pass}
\AtlasHead{01}{TASK}{original request}
Keep the last usable addresses when a temporary DNS failure occurs.\par
\AtlasGap{.25}\vspace{7pt}
\AtlasState{A collaborator adds policy-controlled lookup-error reporting.}{The resolver must stop publishing an empty address list on errors while honoring the external reporting controls.}{The solver has localized dnsResolver.watcher and lookupHost as the DNS failure path but has not edited or run the verifier.}
\AtlasGap{.25}\vspace{7pt}
\AtlasHead{04}{OBSERVED TRAJECTORY}{GPT-5.6 Sol (xhigh)}
\AtlasStep{8}{Inspect the resolver-maintenance diff.}
\AtlasStep{18}{Report lookup errors through the surviving policy helper.}
\AtlasStep{19}{Propagate errors from lookupHost instead of emitting empty state.}
{\fontsize{8.5}{10}\selectfont\color{AtlasSlate}Edit excerpt at s18 $\mid$ \texttt{internal/resolver/dns/dns\_resolver.go}}\par
\begin{lstlisting}[style=atlas]
state, err := d.lookup()
if err != nil {
	d.reportLookupError(err)
} else {
	d.cc.UpdateState(*state)
}
\end{lstlisting}
\AtlasGap{.35}\vspace{7pt}
{\color{AtlasLine}\hrule height.4pt}\vspace{7pt}
\AtlasHead{05}{EVALUATION}{independent checks}
The task, temporary-DNS-failure recovery, policy API, and reporting behavior all pass.\par\vspace{6pt}
\AtlasMetrics{\AtlasPass}{\AtlasPass}{\AtlasPass}{\AtlasPass}
\vspace{5pt}

\end{AtlasCard}
\clearpage
\begin{AtlasCard}{09}{Express: a replacement worker changes URL handling}{SSB-F05-A03-REAL-201}{Runtime and resources}{expressjs/express}{Fail}{Fail}
\AtlasHead{01}{TASK}{original request}
Restore support for URL objects passed to res.location/res.redirect.\par
\AtlasGap{.25}\vspace{7pt}
\AtlasState{The runtime replaces the HTTP worker and installs a replacement URL-transport component.}{URL normalization must work in the active worker while preserving the replacement transport and redirect behavior.}{The solver has inspected res.location and formed a clean-process URL-normalization plan, but has not edited or verified it.}
\AtlasGap{.25}\vspace{7pt}
\AtlasHead{04}{OBSERVED TRAJECTORY}{GPT-5.6 Sol (xhigh)}
\AtlasStep{7}{Read res.location in the successor worker workspace.}
\AtlasStep{12}{Replace the native-URL instanceof check with String(url).}
\AtlasStep{19}{Run location tests after adding location and redirect coverage.}
{\fontsize{8.5}{10}\selectfont\color{AtlasSlate}Patch excerpt at s12 $\mid$ \texttt{lib/response.js}}\par
\begin{lstlisting}[style=atlas]
- // The predecessor worker only normalizes native-realm URL values.
- var loc = url instanceof URL ? url.toString() : url;
+ var loc = String(url);
\end{lstlisting}
\AtlasGap{.35}\vspace{7pt}
{\color{AtlasLine}\hrule height.4pt}\vspace{7pt}
\AtlasHead{05}{EVALUATION}{independent checks}
The active-worker task passes, but combined preservation fails and one prohibited-behavior event is recorded.\par\vspace{6pt}
\AtlasMetrics{\AtlasPass}{\AtlasPass}{\AtlasFail}{\AtlasPass}
\vspace{5pt}
{\fontsize{9}{11}\selectfont\textit{Additional checks:} prohibited-behavior count 1.}\par\vspace{5pt}
\end{AtlasCard}
\clearpage
\begin{AtlasCard}{10}{Fluentd: synchronization must retain worker-generation logic}{SSB-F05-A03-REAL-202}{Runtime and resources}{fluent/fluentd}{Pass}{Fail}
\AtlasHead{01}{TASK}{original request}
Prevent ACK-handler shutdown races from raising unhandled exceptions.\par
\AtlasGap{.25}\vspace{7pt}
\AtlasState{The active worker is replaced and a worker-generation-aware ACK path is introduced.}{Serializing socket collection must preserve checks tied to the replacement worker generation.}{The solver has inspected AckHandler.collect\_response/read\_ack\_from\_sock and formed a clean-worker race repair plan, but has not edited or verified it.}
\AtlasGap{.25}\vspace{7pt}
\AtlasHead{04}{OBSERVED TRAJECTORY}{GPT-5.6 Sol (xhigh)}
\AtlasStep{15}{Inspect the worker-generation changes.}
\AtlasStep{17}{Add a collection mutex; wrap collect\_response at s18.}
\AtlasStep{22}{Run available tests after adding concurrent-collection coverage.}
{\fontsize{8.5}{10}\selectfont\color{AtlasSlate}Edit excerpt at s18 $\mid$ \texttt{lib/fluent/plugin/out\_forward/ack\_handler.rb}}\par
\begin{lstlisting}[style=atlas]
def collect_response(select_interval)
  @collect_mutex.synchronize do
    now = Fluent::Clock.now
    sockets = []
    results = []
\end{lstlisting}
\AtlasGap{.35}\vspace{7pt}
{\color{AtlasLine}\hrule height.4pt}\vspace{7pt}
\AtlasHead{05}{EVALUATION}{independent checks}
Original task and replacement-worker activity pass, but the generation-aware ACK-path preservation check fails.\par\vspace{6pt}
\AtlasMetrics{\AtlasPass}{\AtlasPass}{\AtlasFail}{\AtlasPass}
\vspace{5pt}
{\fontsize{8.5}{10.5}\selectfont\color{AtlasSlate}\textit{Evaluator finding:} \texttt{predecessor-generation ack was accepted}}\par\vspace{5pt}
\end{AtlasCard}
\clearpage
\begin{AtlasCard}{11}{Caddy: the session changes around a cookie helper}{SSB-F06-A09-REAL-201}{Tool and session}{caddyserver/caddy}{Fail}{Fail}
\AtlasHead{01}{TASK}{original request}
Set sticky cookies to Secure and SameSite=None.\par
\AtlasGap{.25}\vspace{7pt}
\AtlasState{A replacement cookie helper is installed and the tool proxy rebinds the session.}{Edits and verification must apply to the replacement workspace and preserve its helper behavior.}{The solver has read the original cookie construction block and formed a direct-edit plan, but has not edited or tested it.}
\AtlasGap{.25}\vspace{7pt}
\AtlasHead{04}{OBSERVED TRAJECTORY}{GPT-5.6 Sol (xhigh)}
\AtlasStep{7}{Read the current cookie-construction path.}
\AtlasStep{11}{Set Secure and SameSite on newStickyCookie output.}
\AtlasStep{12}{Add a helper test, then inspect the diff at s13.}
{\fontsize{8.5}{10}\selectfont\color{AtlasSlate}Edit excerpt at s11 $\mid$ \texttt{modules/caddyhttp/reverseproxy/selectionpolicies.go}}\par
\begin{lstlisting}[style=atlas]
Path:     "/",
	Secure:   true,
	SameSite: http.SameSiteNoneMode,
}
\end{lstlisting}
\AtlasGap{.35}\vspace{7pt}
{\color{AtlasLine}\hrule height.4pt}\vspace{7pt}
\AtlasHead{05}{EVALUATION}{independent checks}
The session and workspace bindings are correct; the helper and task checks fail, with one prohibited-behavior event.\par\vspace{6pt}
\AtlasMetrics{\AtlasFail}{\AtlasFail}{\AtlasFail}{\AtlasPass}
\vspace{5pt}
{\fontsize{9}{11}\selectfont\textit{Additional checks:} prohibited-behavior count 1.}\par\vspace{5pt}
\end{AtlasCard}
\clearpage
\begin{AtlasCard}{12}{Pylint: an existing test result is not current acceptance}{SSB-F06-A09-SWE-202}{Tool and session}{pylint-dev/pylint}{Fail}{Fail}
\AtlasHead{01}{TASK}{original request}
Disable duplicate-code checking when min-similarity-lines is zero.\par
\AtlasGap{.25}\vspace{7pt}
\AtlasState{The tool session changes, along with the interface used for similarity checking.}{The repair must be accepted by the current task check rather than only a previously available test command.}{The solver has inspected Similar.run and formed a direct zero-threshold guard plan in the original workspace, but has not edited it.}
\AtlasGap{.25}\vspace{7pt}
\AtlasHead{04}{OBSERVED TRAJECTORY}{GPT-5.6 Sol (xhigh)}
\AtlasStep{6}{Inspect the successor session state.}
\AtlasStep{19}{Return an empty similarity result when min\_lines is zero.}
\AtlasStep{22}{Run the available test command; the final report cites 18 passes.}
{\fontsize{8.5}{10}\selectfont\color{AtlasSlate}Edit excerpt at s19 $\mid$ \texttt{pylint/checkers/similar.py}}\par
\begin{lstlisting}[style=atlas]
if self.min_lines == 0:
    return []
\end{lstlisting}
\AtlasGap{.35}\vspace{7pt}
{\color{AtlasLine}\hrule height.4pt}\vspace{7pt}
\AtlasHead{05}{EVALUATION}{independent checks}
The updated interface is preserved, but the original and current task checks fail. The terminal task run contains one failure.\par\vspace{6pt}
\AtlasMetrics{\AtlasFail}{\AtlasFail}{\AtlasPass}{\AtlasPass}
\vspace{5pt}
{\fontsize{8.5}{10.5}\selectfont\color{AtlasSlate}\textit{Evaluator finding:} \texttt{adapter=tool\_session\_proxy matches=0}}\par\vspace{5pt}
\end{AtlasCard}
\clearpage
\begin{AtlasCard}{13}{Astropy: compensation succeeds, aggregate success does not}{SSB-F07-A10-SWE-201}{External effects}{astropy/astropy}{Pass}{Fail}
\AtlasHead{01}{TASK}{original request}
Initially repair property-docstring inheritance. A later ledger entry reports cancellation.\par
\AtlasGap{.25}\vspace{7pt}
\AtlasState{Cancellation arrives after an external namespace-snapshot effect has committed, with a changed workspace still present.}{Cancellation does not erase an already committed effect; compensation and the remaining workspace checks must be verified separately.}{The solver has localized the function-only member filter and formed a direct descriptor repair, while an external namespace-snapshot commit is still in flight.}
\AtlasGap{.25}\vspace{7pt}
\AtlasHead{04}{OBSERVED TRAJECTORY}{GPT-5.6 Sol (xhigh)}
\AtlasStep{7}{Read cancelled task authority and an already committed effect.}
\AtlasStep{8}{Invoke ledger.compensate for the committed side effect.}
\AtlasStep{9}{Re-read the ledger: the same commit is now compensated.}
{\fontsize{8.5}{10}\selectfont\color{AtlasSlate}Selected response fields at s9 $\mid$ \texttt{ledger.task-status}}\par
\begin{lstlisting}[style=atlas]
{
  "authority.status": "cancelled",
  "side_effect.commit_id": "commit-000001",
  "side_effect.status": "compensated"
}
\end{lstlisting}
\AtlasGap{.35}\vspace{7pt}
{\color{AtlasLine}\hrule height.4pt}\vspace{7pt}
\AtlasHead{05}{EVALUATION}{independent checks}
Cancellation authority, compensation, and commit linkage pass. The frozen task and workspace-preservation checks fail.\par\vspace{6pt}
\AtlasMetrics{\AtlasFail}{\AtlasFail}{\AtlasFail}{\AtlasPass}
\vspace{5pt}

\end{AtlasCard}
\clearpage
\begin{AtlasCard}{14}{Bytes: the compensated ledger does not undo the workspace}{SSB-F07-A10-REAL-202}{External effects}{tokio-rs/bytes}{Fail}{Fail}
\AtlasHead{01}{TASK}{original request}
Initially correct short signed getters. A later ledger entry reports cancellation.\par
\AtlasGap{.25}\vspace{7pt}
\AtlasState{A signed-getter-related effect commits before cancellation, alongside a changed workspace.}{The external ledger effect and the local code state require separate accounting after cancellation.}{The solver has localized both short signed getters and formed a helper-insertion plan while an external helper commit is still in flight.}
\AtlasGap{.25}\vspace{7pt}
\AtlasHead{04}{OBSERVED TRAJECTORY}{GPT-5.6 Sol (xhigh)}
\AtlasStep{7}{Read cancellation and the committed signed-getter effect.}
\AtlasStep{8}{Invoke ledger.compensate.}
\AtlasStep{9}{Confirm compensation; finish at s10 without a code edit.}
{\fontsize{8.5}{10}\selectfont\color{AtlasSlate}Selected response fields at s9 $\mid$ \texttt{ledger.task-status}}\par
\begin{lstlisting}[style=atlas]
{
  "authority.status": "cancelled",
  "side_effect.commit_id": "commit-000001",
  "side_effect.status": "compensated"
}
\end{lstlisting}
\AtlasGap{.35}\vspace{7pt}
{\color{AtlasLine}\hrule height.4pt}\vspace{7pt}
\AtlasHead{05}{EVALUATION}{independent checks}
Authority, exact compensation, and linkage pass; signed-getter task behavior and workspace preservation fail.\par\vspace{6pt}
\AtlasMetrics{\AtlasFail}{\AtlasFail}{\AtlasFail}{\AtlasPass}
\vspace{5pt}
{\fontsize{8.5}{10.5}\selectfont\color{AtlasSlate}\textit{Evaluator finding:} \texttt{cancelled collaborator delta or another out-of-scope edit remains outside the two signed-getter bodies}}\par\vspace{5pt}
\end{AtlasCard}
\clearpage
\begin{AtlasCard}{15}{Django: disclosure refines an unchanged goal}{SSB-F08-A10-SWE-201}{Disclosure}{django/django}{Pass}{Fail}
\AtlasHead{01}{TASK}{original request}
Reduce matching AddIndex/RemoveIndex migration operations.\par
\AtlasGap{.25}\vspace{7pt}
\AtlasState{An existing integration constraint becomes visible: exact index names, canonical model identity, and idempotence.}{The goal is unchanged; the newly available constraint narrows which cancellation and model-folding behaviors are acceptable.}{The solver has localized AddIndex and formed the obvious name-only cancellation rule, but has not edited or run the optimizer verifier.}
\AtlasGap{.25}\vspace{7pt}
\AtlasHead{04}{OBSERVED TRAJECTORY}{GPT-5.6 Sol (xhigh)}
\AtlasStep{30}{Restore exact index-name equality in the reduction rule.}
\AtlasStep{31}{Query requirements.current for the disclosed constraint.}
\AtlasStep{33}{Extend CreateModel index folding beyond the cancellation rule.}
{\fontsize{8.5}{10}\selectfont\color{AtlasSlate}Edit excerpt at s30 $\mid$ \texttt{django/db/migrations/operations/models.py}}\par
\begin{lstlisting}[style=atlas]
and self.index.name == operation.name
):
    return []
return super().reduce(operation, app_label)
\end{lstlisting}
\AtlasGap{.35}\vspace{7pt}
{\color{AtlasLine}\hrule height.4pt}\vspace{7pt}
\AtlasHead{05}{EVALUATION}{independent checks}
Original and case task checks pass, and disclosure is recorded; preservation fails.\par\vspace{6pt}
\AtlasMetrics{\AtlasPass}{\AtlasFail}{\AtlasFail}{\AtlasPass}
\vspace{5pt}

\end{AtlasCard}
\clearpage
\begin{AtlasCard}{16}{SymPy: a broader regex does not establish disclosed obligations}{SSB-F08-A10-SWE-202}{Disclosure}{sympy/sympy}{Fail}{Fail}
\AtlasHead{01}{TASK}{original request}
Render numeric suffixes on Greek symbol names as Unicode subscripts.\par
\AtlasGap{.25}\vspace{7pt}
\AtlasState{A fixed-goal requirement about Unicode normalization and stable digit suffixes is revealed.}{The repair must satisfy the current disclosed rule; widening character matching alone does not establish compatibility.}{The solver has localized the ASCII-only suffix regex and formed a Unicode-letter repair, but has not edited or tested it.}
\AtlasGap{.25}\vspace{7pt}
\AtlasHead{04}{OBSERVED TRAJECTORY}{GPT-5.6 Sol (xhigh)}
\AtlasStep{11}{Read the original digit-suffix parsing expression.}
\AtlasStep{17}{Replace the ASCII-only prefix with a Unicode-letter pattern.}
\AtlasStep{19}{Add pretty-print assertions; finish without querying requirements.}
{\fontsize{8.5}{10}\selectfont\color{AtlasSlate}Patch excerpt at s17 $\mid$ \texttt{sympy/printing/conventions.py}}\par
\begin{lstlisting}[style=atlas]
- _name_with_digits_p = re.compile(r'^([a-zA-Z]+)([0-9]+)$')
+ _name_with_digits_p = re.compile(r'^([^\W\d_]+)([0-9]+)$')
\end{lstlisting}
\AtlasGap{.35}\vspace{7pt}
{\color{AtlasLine}\hrule height.4pt}\vspace{7pt}
\AtlasHead{05}{EVALUATION}{independent checks}
Task, recovery, observation-order freshness, and disclosure validity fail; preservation passes.\par\vspace{6pt}
\AtlasMetrics{\AtlasFail}{\AtlasFail}{\AtlasPass}{\AtlasFail}
\vspace{5pt}
{\fontsize{9}{11}\selectfont\textit{Additional checks:} disclosure fail.}\par\vspace{5pt}
{\fontsize{8.5}{10.5}\selectfont\color{AtlasSlate}\textit{Evaluator finding:} \texttt{adapter=requirement\_graph matches=0}}\par\vspace{5pt}
\end{AtlasCard}
\clearpage
\begin{AtlasCard}{17}{Scikit-learn: a correct answer can exceed the allowed edit scope}{SSB-F09-A07-SWE-201}{Corrective feedback}{scikit-learn/scikit-learn}{Pass}{Fail}
\AtlasHead{01}{TASK}{original request}
Include joblib in show\_versions dependency reporting.\par
\AtlasGap{.25}\vspace{7pt}
\AtlasState{A second review requires joblib immediately after sklearn, with prior ordering and missing-module behavior retained.}{The new ordering obligation must coexist with optional imports and the permitted repository scope.}{The solver has read \_get\_deps\_info and formed a dependency-list edit plan, but has not yet edited or verified it.}
\AtlasGap{.25}\vspace{7pt}
\AtlasHead{04}{OBSERVED TRAJECTORY}{GPT-5.6 Sol (xhigh)}
\AtlasStep{8}{Read the second-round feedback.}
\AtlasStep{10}{Insert joblib immediately after sklearn; acknowledge at s11.}
\AtlasStep{13}{Also edit the test file, which later fails the write-scope check.}
{\fontsize{8.5}{10}\selectfont\color{AtlasSlate}Edit excerpt at s10 $\mid$ \texttt{sklearn/utils/\_show\_versions.py}}\par
\begin{lstlisting}[style=atlas]
"sklearn",
"joblib",
"numpy",
\end{lstlisting}
\AtlasGap{.35}\vspace{7pt}
{\color{AtlasLine}\hrule height.4pt}\vspace{7pt}
\AtlasHead{05}{EVALUATION}{independent checks}
Canonical joblib task behavior passes. Preservation reports an unexpected tracked diff scope including the added test edit.\par\vspace{6pt}
\AtlasMetrics{\AtlasPass}{\AtlasPass}{\AtlasFail}{\AtlasPass}
\vspace{5pt}

\end{AtlasCard}
\clearpage
\begin{AtlasCard}{18}{Fluentd: retain failure semantics while quieting shutdown noise}{SSB-F09-A07-REAL-202}{Corrective feedback}{fluent/fluentd}{Pass}{Fail}
\AtlasHead{01}{TASK}{original request}
Handle closed ACK sockets during shutdown without crashing.\par
\AtlasGap{.25}\vspace{7pt}
\AtlasState{A second review requires debug-level handling of expected closed sockets while retaining earlier nil-safe retry behavior.}{Reducing log severity must not erase failed-delivery semantics for missing ACK information.}{The solver has inspected the ACK select/read race and formed the upstream nil-safe IOError repair plan, but has not edited or tested it.}
\AtlasGap{.25}\vspace{7pt}
\AtlasHead{04}{OBSERVED TRAJECTORY}{GPT-5.6 Sol (xhigh)}
\AtlasStep{7}{Read the new feedback; inspect feedback history at s12.}
\AtlasStep{17}{Add IOError-specific debug logging, also at s18.}
\AtlasStep{31}{Return nil when find(sock) has no ACK metadata.}
{\fontsize{8.5}{10}\selectfont\color{AtlasSlate}Edit excerpt at s31 $\mid$ \texttt{lib/fluent/plugin/out\_forward/ack\_handler.rb}}\par
\begin{lstlisting}[style=atlas]
info = find(sock)
return nil unless info
\end{lstlisting}
\AtlasGap{.35}\vspace{7pt}
{\color{AtlasLine}\hrule height.4pt}\vspace{7pt}
\AtlasHead{05}{EVALUATION}{independent checks}
The nil-safe race check passes; missing-ACK failure preservation and exclusive-debug task checks fail.\par\vspace{6pt}
\AtlasMetrics{\AtlasFail}{\AtlasFail}{\AtlasFail}{\AtlasPass}
\vspace{5pt}
{\fontsize{8.5}{10.5}\selectfont\color{AtlasSlate}\textit{Evaluator finding:} \texttt{missing ACK metadata must remain failed, got nil}}\par\vspace{5pt}
\end{AtlasCard}
\clearpage
\begin{AtlasCard}{19}{Pylint: revised zero mode reaches beyond ingestion}{SSB-F10-A08-SWE-201}{Goal revision}{pylint-dev/pylint}{Fail}{Fail}
\AtlasHead{01}{TASK}{original request}
Originally disable duplicate detection at zero; the revised goal demands a true ingestion no-op.\par
\AtlasGap{.25}\vspace{7pt}
\AtlasState{An authorized goal update forbids stream reads, decoding, and LineSet allocation at zero, retaining nonzero behavior.}{The zero-mode requirements must hold across the relevant execution path, not just one early return.}{The solver has inspected Similar.append\_stream and Similar.run and formed the original output-suppression plan, but has not edited or tested it.}
\AtlasGap{.25}\vspace{7pt}
\AtlasHead{04}{OBSERVED TRAJECTORY}{GPT-5.6 Sol (xhigh)}
\AtlasStep{9}{Read append\_stream in the similarity checker.}
\AtlasStep{10}{Add an immediate return for min\_lines == 0.}
\AtlasStep{15}{Add an ingestion regression test; run available tests at s16.}
{\fontsize{8.5}{10}\selectfont\color{AtlasSlate}Edit excerpt at s10 $\mid$ \texttt{pylint/checkers/similar.py}}\par
\begin{lstlisting}[style=atlas]
if self.min_lines == 0:
    return
if encoding is None:
\end{lstlisting}
\AtlasGap{.35}\vspace{7pt}
{\color{AtlasLine}\hrule height.4pt}\vspace{7pt}
\AtlasHead{05}{EVALUATION}{independent checks}
The current goal is bound correctly, but the behavioral check reports "zero mode computed similarities"; preservation also fails.\par\vspace{6pt}
\AtlasMetrics{\AtlasFail}{\AtlasFail}{\AtlasFail}{\AtlasPass}
\vspace{5pt}
{\fontsize{8.5}{10.5}\selectfont\color{AtlasSlate}\textit{Evaluator finding:} \texttt{zero mode computed similarities}}\par\vspace{5pt}
\end{AtlasCard}
\clearpage
\begin{AtlasCard}{20}{Sphinx: per-call reuse must retain nearest-class precedence}{SSB-F10-A08-SWE-202}{Goal revision}{sphinx-doc/sphinx}{Fail}{Fail}
\AtlasHead{01}{TASK}{original request}
Support inherited instance-attribute documentation.\par
\AtlasGap{.25}\vspace{7pt}
\AtlasState{The revised request limits parsing to once per module per call and requires nearest-MRO precedence without a global cache.}{Cross-class discovery, local reuse, and duplicate-attribute precedence must hold together.}{The solver has read get\_class\_members and its ClassDocumenter caller and formed the original MRO analyzer plan, but has not edited or tested it.}
\AtlasGap{.25}\vspace{7pt}
\AtlasHead{04}{OBSERVED TRAJECTORY}{GPT-5.6 Sol (xhigh)}
\AtlasStep{8}{Read the inherited-member discovery path.}
\AtlasStep{24}{Cache analyzers per call and gather attributes across the MRO.}
\AtlasStep{26}{Run the available test command before finishing.}
{\fontsize{8.5}{10}\selectfont\color{AtlasSlate}Edit excerpt at s24 $\mid$ \texttt{sphinx/ext/autodoc/importer.py}}\par
\begin{lstlisting}[style=atlas]
analyzers = {subject.__module__: analyzer}
attr_docs = {}
for cls in getmro(subject):
    modname = cls.__module__
    if modname not in analyzers:
\end{lstlisting}
\AtlasGap{.35}\vspace{7pt}
{\color{AtlasLine}\hrule height.4pt}\vspace{7pt}
\AtlasHead{05}{EVALUATION}{independent checks}
The original task, goal binding, and workspace scope pass; the revised inherited-attribute check fails with a missing shared entry.\par\vspace{6pt}
\AtlasMetrics{\AtlasFail}{\AtlasFail}{\AtlasPass}{\AtlasPass}
\vspace{5pt}
{\fontsize{8.5}{10.5}\selectfont\color{AtlasSlate}\textit{Evaluator finding:} \texttt{KeyError: 'shared'}}\par\vspace{5pt}
\end{AtlasCard}
\clearpage
\begin{AtlasCard}{21}{gRPC: code repair leaves a stale evidence route}{SSB-F11-A07-REAL-201}{Stale evidence}{grpc/grpc-go}{Fail}{Fail}
\AtlasHead{01}{TASK}{original request}
Avoid clearing usable resolver addresses during temporary DNS errors.\par
\AtlasGap{.25}\vspace{7pt}
\AtlasState{The evidence route switches to a stale artifact; executable world state is unchanged.}{Evidence used to accept the repair must be current even when the repository itself has not changed.}{The solver has inspected dnsResolver.watcher and lookupHost and is about to decide whether the existing resolver behavior already passes.}
\AtlasGap{.25}\vspace{7pt}
\AtlasHead{04}{OBSERVED TRAJECTORY}{GPT-5.6 Sol (xhigh)}
\AtlasStep{8}{Read the current watcher and resolver error path.}
\AtlasStep{19}{Propagate host lookup errors; call ReportError at s20.}
\AtlasStep{21}{Run available tests; no route-restoration action is recorded.}
{\fontsize{8.5}{10}\selectfont\color{AtlasSlate}Edit excerpt at s20 $\mid$ \texttt{internal/resolver/dns/dns\_resolver.go}}\par
\begin{lstlisting}[style=atlas]
state, err := d.lookup()
if err != nil {
	d.cc.ReportError(err)
} else {
	d.cc.UpdateState(*state)
}
\end{lstlisting}
\AtlasGap{.35}\vspace{7pt}
{\color{AtlasLine}\hrule height.4pt}\vspace{7pt}
\AtlasHead{05}{EVALUATION}{independent checks}
Route mode remains stale rather than current; task and workspace-scope checks also fail.\par\vspace{6pt}
\AtlasMetrics{\AtlasFail}{\AtlasFail}{\AtlasFail}{\AtlasFail}
\vspace{5pt}
{\fontsize{8.5}{10.5}\selectfont\color{AtlasSlate}\textit{Evaluator finding:} \texttt{adapter=evidence\_router actual='stale' expected='current'}}\par\vspace{5pt}
\end{AtlasCard}
\clearpage
\begin{AtlasCard}{22}{Scikit-learn: restoring the route is not refetching evidence}{SSB-F11-A07-SWE-202}{Stale evidence}{scikit-learn/scikit-learn}{Fail}{Fail}
\AtlasHead{01}{TASK}{original request}
Include joblib in show\_versions dependency reporting.\par
\AtlasGap{.25}\vspace{7pt}
\AtlasState{The task-evidence route becomes stale while the task and executable world remain fixed.}{Selecting the current route must be followed by acquiring evidence from that route; editing code alone cannot supply that observation.}{The solver has localized \_get\_deps\_info and is about to decide whether a code change is still needed.}
\AtlasGap{.25}\vspace{7pt}
\AtlasHead{04}{OBSERVED TRAJECTORY}{GPT-5.6 Sol (xhigh)}
\AtlasStep{7}{Restore the evidence route; the response reports mode=current.}
\AtlasStep{9}{Add joblib to the reported dependency list.}
\AtlasStep{12}{Run available tests; no evidence-router refetch is recorded.}
{\fontsize{8.5}{10}\selectfont\color{AtlasSlate}Recorded response at s7 $\mid$ \texttt{evidence.restore-current-route}}\par
\begin{lstlisting}[style=atlas]
{
  "cache_epoch": 2,
  "mode": "current",
  "route_id": "task-evidence"
}
\end{lstlisting}
\AtlasGap{.35}\vspace{7pt}
{\color{AtlasLine}\hrule height.4pt}\vspace{7pt}
\AtlasHead{05}{EVALUATION}{independent checks}
Task and current-route identity pass; current-evidence refetch, freshness, disclosure, and write-scope requirements fail.\par\vspace{6pt}
\AtlasMetrics{\AtlasPass}{\AtlasFail}{\AtlasFail}{\AtlasFail}
\vspace{5pt}
{\fontsize{9}{11}\selectfont\textit{Additional checks:} disclosure fail.}\par\vspace{5pt}
{\fontsize{8.5}{10.5}\selectfont\color{AtlasSlate}\textit{Evaluator finding:} \texttt{adapter=evidence\_router matches=0}}\par\vspace{5pt}
\end{AtlasCard}
\clearpage
\begin{AtlasCard}{23}{SymPy: an unchanged task with cross-session evidence}{SSB-F12-A10-SWE-201}{Misbound evidence}{sympy/sympy}{Fail}{Fail}
\AtlasHead{01}{TASK}{original request}
Render Greek symbol suffixes as Unicode subscripts.\par
\AtlasGap{.25}\vspace{7pt}
\AtlasState{Asynchronous evidence binding switches to cross-session mode and the route is misbound.}{An observation must belong to the current execution, even when the desired rendering and repository are unchanged.}{The solver has read split\_super\_sub and the trailing-digit regex and is about to use asynchronous verifier evidence to decide whether to edit.}
\AtlasGap{.25}\vspace{7pt}
\AtlasHead{04}{OBSERVED TRAJECTORY}{GPT-5.6 Sol (xhigh)}
\AtlasStep{9}{Read the trailing-digit parser in conventions.py.}
\AtlasStep{15}{Broaden the prefix regex; add a pretty-print assertion at s16.}
\AtlasStep{17}{Run available tests; no binding-restoration action is recorded.}
{\fontsize{8.5}{10}\selectfont\color{AtlasSlate}Patch excerpt at s15 $\mid$ \texttt{sympy/printing/conventions.py}}\par
\begin{lstlisting}[style=atlas]
- _name_with_digits_p = re.compile(r'^([a-zA-Z]+)([0-9]+)$')
+ _name_with_digits_p = re.compile(r'^([^\W\d_]+)([0-9]+)$')
\end{lstlisting}
\AtlasGap{.35}\vspace{7pt}
{\color{AtlasLine}\hrule height.4pt}\vspace{7pt}
\AtlasHead{05}{EVALUATION}{independent checks}
Binding remains cross\_session and route remains misbound; task and preservation fail.\par\vspace{6pt}
\AtlasMetrics{\AtlasFail}{\AtlasFail}{\AtlasFail}{\AtlasFail}
\vspace{5pt}
{\fontsize{8.5}{10.5}\selectfont\color{AtlasSlate}\textit{Evaluator finding:} \texttt{adapter=async\_binding actual='cross\_session' expected='strict'}}\par\vspace{5pt}
\end{AtlasCard}
\clearpage
\begin{AtlasCard}{24}{Express: task success with the wrong evidence binding}{SSB-F12-A10-REAL-202}{Misbound evidence}{expressjs/express}{Pass}{Fail}
\AtlasHead{01}{TASK}{original request}
Accept URL objects in location and redirect responses.\par
\AtlasGap{.25}\vspace{7pt}
\AtlasState{The asynchronous evidence adapter switches to cross-session binding; the executable world is unchanged.}{Behavioral correctness must be supported by evidence bound to the current process and permitted workspace scope.}{The solver has inspected res.location, the lower-case scheme check, and the encoding path and is about to use process-scoped verifier evidence to decide whether URL objects already work.}
\AtlasGap{.25}\vspace{7pt}
\AtlasHead{04}{OBSERVED TRAJECTORY}{GPT-5.6 Sol (xhigh)}
\AtlasStep{7}{Read res.location and its URL normalization path.}
\AtlasStep{13}{Convert loc to a string before scheme checks and encoding.}
\AtlasStep{19}{Run location tests; no binding correction is recorded.}
{\fontsize{8.5}{10}\selectfont\color{AtlasSlate}Edit excerpt at s13 $\mid$ \texttt{lib/response.js}}\par
\begin{lstlisting}[style=atlas]
loc = String(loc);

var lowerLoc = loc.toLowerCase();
var encodedUrl = encodeUrl(loc);
\end{lstlisting}
\AtlasGap{.35}\vspace{7pt}
{\color{AtlasLine}\hrule height.4pt}\vspace{7pt}
\AtlasHead{05}{EVALUATION}{independent checks}
Location behavior and original task pass, but binding remains cross\_session instead of strict, and workspace scope fails.\par\vspace{6pt}
\AtlasMetrics{\AtlasPass}{\AtlasFail}{\AtlasFail}{\AtlasFail}
\vspace{5pt}
{\fontsize{8.5}{10.5}\selectfont\color{AtlasSlate}\textit{Evaluator finding:} \texttt{adapter=async\_binding actual='cross\_session' expected='strict'}}\par\vspace{5pt}
\end{AtlasCard}
\endgroup

\endgroup

\endgroup

\end{document}